\documentclass{article}

\newcommand{\PaperRoot}{.}

\usepackage{microtype}
\usepackage{graphicx}
\usepackage{capt-of}
\usepackage{subfigure}
\usepackage{booktabs}
\usepackage{array}
\usepackage{comment}
\usepackage{placeins}
\usepackage{hyperref}

\usepackage[accepted]{icml2026}

\usepackage{amsmath}
\usepackage{amssymb}
\usepackage{mathtools}
\usepackage{amsthm}

\DeclareMathOperator{\ReLU}{ReLU}
\newcommand{\relu}{\ensuremath{\ReLU}}

\newcolumntype{L}[1]{>{\raggedright\arraybackslash}p{#1}}

\usepackage[capitalize,noabbrev]{cleveref}

\theoremstyle{plain}

\theoremstyle{definition}

\theoremstyle{remark}

\icmltitlerunning{Dropout Universality: Scaling Laws and Optimal Scheduling at the Edge-of-Chaos}

\begin{document}

\twocolumn[
 \icmltitle{Dropout Universality: Scaling Laws and Optimal Scheduling at the Edge-of-Chaos}

 \icmlsetsymbol{equal}{*}

 \begin{icmlauthorlist}
 \icmlauthor{Lucas Fernandez Sarmiento}{cmu}
 \end{icmlauthorlist}

 \icmlaffiliation{cmu}{Carnegie Mellon University (CMU), Pittsburgh, Pennsylvania, USA}

 \icmlcorrespondingauthor{Lucas Fernandez Sarmiento}{lucasf@andrew.cmu.edu}

 \icmlkeywords{dropout, mean-field theory, edge of chaos, critical scaling, neural network initialization}

 \vskip 0.3in
]

\printAffiliationsAndNotice{}

\begin{abstract}
We ask whether the standard treatment of dropout as a static hyperparameter is optimal, or whether its utility can be improved by letting it vary over depth. We answer this by developing a mean-field theory of dropout near the edge of chaos, identifying distinct universality classes for smooth and kinked activations, together with their scaling exponents. The resulting propagation theory, constrained by maximizing the regularization delivered by dropout, motivates concentrating dropout near the input. Experiments on vision, speech and financial time series show gains most consistently in MLPs, with smaller gains in Transformers. 
\end{abstract}

\section{Introduction}
Dropout is used to ``corrupt'' a network so that it relies less on particular paths from input to output \cite{srivastava2014dropout}. Keeping the average dropout fixed, where should we introduce this corruption to make it most useful? We study this question near the edge of chaos, asking whether redistributing dropout across layers can preserve information propagation while exposing more of the network to regularization \cite{poole2016exponentialexpressivitydeepneural,schoenholz2017deepinformationpropagation,Bahri2020StatMechDeepLearning,bahri2024leshoucheslecturesdeep}.

In standard mean-field theory, driving a neural network to criticality means delaying the approach to perfect correlation that can result from passing distinct inputs through an identical network. Indistinguishable representations at the output layer leave no information to tell the inputs apart, and thus, avoiding this is paramount to ensure proper performance in neural networks. However, the presence of dropout causes the perfect-correlation fixed point to evaporate \cite{schoenholz2017deepinformationpropagation}, as even identical inputs see different representations due to the independent masks blocking a percentage of activations at a given layer. Correlation lengths derived from mean-field theory serve as an effective decay scale, much like a half-life for information, and are instrumental in model building, as networks much deeper than this length can be difficult to train. But if dropout removes the perfect-correlation fixed point, where do correlations settle instead, and how many layers does it take to get there? In this work, we derive what this fixed point is using a mean-field theory computation perturbative in dropout, and use it to determine the effective correlation length, showing that this actually depends on a combination of distance to criticality and dropout.

In the representation-group (coarse-graining) language of \cite{roberts2022principlesdeeplearningtheory,bahri2024leshoucheslecturesdeep}, dropout is a relevant perturbation: its effect grows under depth coarse-graining and moves the dynamics away from criticality. The depth correlation length is then finite even at the clean network's edge of chaos \cite{sompolinsky1988chaos,packard1988adaptation,bertschinger2004real}. Dropout can reduce overfitting while shortening the distance over which information propagates, and we use the propagation theory to guide where it is applied, then test whether this improves performance over uniform dropout.

Expanding the correlation map near perfect alignment gives an equation for the decorrelation $m \equiv 1-c^\ast$, where $c^\ast$ is the asymptotic inter-signal correlation. The leading nonlinear term depends on the activation: smooth and kinked (\relu{}-like) activations have the same qualitative phase diagram but different power laws in the presence of dropout.

We also show that changing dropout strength and distance to criticality gives curves that collapse onto a universal form within each activation class, and compare this prediction with mean-field recursions.

While dropout was previously shown to destroy the order-to-chaos critical point \cite{schoenholz2017deepinformationpropagation}, we show that the deformed mean-field map still retains a nontrivial $c^\ast<1$ fixed point. This allows us to have a well-defined correlation length that can guide architecture depth. We show that the analytic structure of activations sets the universality classes and the scaling laws for information propagation: smooth and kinked activations have distinct critical exponents (Table~\ref{tab:critical_exponents}), a distinction reflected in their Hermite spectral structure (App.~\ref{sec:var-opt-act}). Since the same Gaussian activation kernels arise beyond MLPs, including CNNs and residual branches (App.~\ref{app:cnn_extension}), we expect the smooth/kinked distinction to help guide these cases too. We also treat dropout as a field that varies with depth, allowing a hyperparameter usually held constant to be optimized across the network, and potentially over training time using dynamical mean-field theory (DMFT) \cite{bordelon2022selfconsistent}.

This work complements prior studies of depth-dependent regularization: stochastic depth drops whole residual blocks with increasing probability \cite{huang2016deep}, curriculum dropout anneals dropout over training time \cite{morerio2017curriculum}, and LayerDrop removes transformer layers for efficiency \cite{fan2019reducing}. We vary dropout across depth, so our schedules can be combined with changes over training time and with the removal of whole layers. Qualitatively different edge-of-chaos behavior for smooth and \relu{}-like activations was previously observed in \cite{hayou2019impact}; here we extract the scaling exponents and identify the corresponding universality classes.\footnote{The notebooks, saved results and confidence interval calculations are available in the \href{https://github.com/luklacasito/icml}{\texttt{icml}} repository. See the \hyperref[sec:reproducibility]{reproducibility statement} for code and data.}

\paragraph{Conflict of Interest Disclosure.}
The author declares no financial conflicts of interest related to this work.

\section{Mean-Field Theory Background}
\label{sec:mft_background}
In this work, we study neural networks at initialization. We assume that the network is wide enough for the central limit theorem to apply, while not deep enough for non-Gaussianities to accumulate to the point where they dominate. This lets us follow the network through the variances and correlations of its activations, rather than tracking every neuron separately \cite{poole2016exponentialexpressivitydeepneural,schoenholz2017deepinformationpropagation,Bahri2020StatMechDeepLearning,bahri2024leshoucheslecturesdeep}.

We consider an MLP of depth $L$, $N_l$ neurons in layer $l$, and an activation $\phi:\mathbb{R}\to\mathbb{R}$ (e.g.\ $\tanh$, \relu{}, \emph{etc.}). We use the standard i.i.d.\ Gaussian initialization
\begin{equation}
W^{l}_{ij}\sim \mathcal{N}\left(0,\frac{\sigma_{w}^{2}}{N_{l-1}}\right), \qquad b^{l}_{i}\sim \mathcal{N}\left(0,\sigma_{b}^{2}\right),
\end{equation}
take the input as $y^{0}_{i;a}=x_{i;a}$, and propagate forward via
\begin{equation}
z^{l}_{i;a}= \sum_{j=1}^{N_{l-1}} W^{l}_{ij} y^{l-1}_{j;a}+b^{l}_{i}, \qquad y^{l}_{i;a}=\phi\left(z^{l}_{i;a}\right),
\end{equation}
where $W^l\in\mathbb R^{N_l\times N_{l-1}}$ and $a$ labels the input. Thus the weight variance is scaled by the number of incoming units. Throughout, expectations $\mathbb{E}[\cdot]$ are over the random weights and biases at initialization, with inputs held fixed. We also use the standard Gaussian measure
\begin{equation}
\int Dz (\cdots)\;\equiv\;\frac{1}{\sqrt{2\pi}}\int_{-\infty}^{\infty}dz e^{-z^{2}/2}(\cdots),
\end{equation}
and similarly for $\int Dz_1 Dz_2$.

At infinite width, the preactivations are Gaussian layer by layer \cite{Lee2018DNNasGP}. With Gaussian initialization and fixed inputs, the first layer is exactly Gaussian even at finite width; later layers acquire non-Gaussian corrections that can accumulate with depth, schematically through $L/N$ \cite{roberts2022principlesdeeplearningtheory,bahri2024leshoucheslecturesdeep}. See App.~\ref{app:mft_primer}.

We track the single-input preactivation variance $q^l_{aa}$, the two-input covariance $q^l_{ab}$, and the induced correlation $c^l_{ab}$. For $l\ge2$, their recursions are
\begin{align}
q^{l}_{aa}&= \sigma_{w}^{2}\int Dz \phi^{2}\left(\sqrt{q^{l-1}_{aa}} z\right)+\sigma_{b}^{2}\\
q^{l}_{ab}&=\sigma_{w}^{2}\int Dz_{1} Dz_{2} \phi(u_{1}) \phi(u_{2})+\sigma_{b}^{2}\\
u_{1}&=\sqrt{q^{l-1}_{aa}} z_{1},\qquad c^{l-1}_{ab}=\frac{q^{l-1}_{ab}}{\sqrt{q^{l-1}_{aa} q^{l-1}_{bb}}},\\
u_{2}&=\sqrt{q^{l-1}_{bb}}\left(c^{l-1}_{ab} z_{1}+\sqrt{1-\left(c^{l-1}_{ab}\right)^{2}} z_{2}\right).
\label{variancerecursion}
\end{align}
The first layer is initialized directly from the inputs, $q^1_{ab}=\sigma_w^2 N_0^{-1}\sum_j x_{j;a}x_{j;b}+\sigma_b^2$.
In the absence of dropout, for a wide array of activations $q^{l}_{ab},c^{l}_{ab}$ settle to fixed points $q^*,c^*$; in particular, there exists a $c^*=1$ fixed point as a straightforward calculation shows. The variance fixed point sets the typical activation scale, while the correlation fixed point describes whether distinct inputs become indistinguishable under depth iteration. We denote the recurrence relation for $c^{l}_{ab}$ by $c^{l}_{ab}=F(c^{l-1}_{ab})$.

To probe its stability, one linearizes the map at $c=1$, defining the (angular) susceptibility
\begin{equation}
\chi_{\perp} \equiv \left.\frac{\partial c^{l}_{ab}}{\partial c^{l-1}_{ab}}\right|_{c_{ab}=1} = \sigma_{w}^{2}\int Dz \big[\phi'(\sqrt{q^{*}} z)\big]^{2}.
\end{equation}
This creates a phase diagram with three regimes: $\chi_{\perp}<1$ is the ordered regime, $\chi_{\perp}>1$ is chaotic, and
 $\chi_{\perp}=1$ defines the critical regime (the so-called "edge-of-chaos"). The depth scale controlling cross-input correlations can become arbitrarily large in the latter, allowing information to penetrate deep into the network. For \relu{}, the criticality condition gives $\sigma_w^2=2$, coinciding with the variance used in He initialization \cite{he2015delvingdeeprectifierssurpassing}\footnote{This is not true for general activations, but rather an idiosyncrasy of \relu{}.} (see App.~\ref{app:mft_primer}), a more fundamental viewpoint than variance preservation.

 This leads to an emergent characteristic depth scale $\xi_c$, which parametrizes both signal propagation and gradient flow. Iterating random affine maps induces an effective coarse-graining in depth akin to renormalization-group flow, called representation group flow in the ML context \cite{roberts2022principlesdeeplearningtheory}; near an attracting correlation fixed point with $r_c=F'(c^*)$ and $0<|r_c|<1$, small correlation deviations decay exponentially with depth,
\begin{equation}
|{c^*-c^l_{ab}}|\propto e^{-l/\xi_{c}}, \qquad \xi_{c}^{-1}\equiv -\log|r_{c}|.
\label{corrq}
\end{equation}
At the marginal point $r_c=1$, the linear term no longer makes deviations shrink, so nonlinear terms determine how they decay, which can follow a power law instead.
An analogous correlation length $\xi_q$ controls how quickly single-input norms settle to $q^{*}$. Heuristically, $\xi_{c}$ controls how far distinctions between different inputs survive with depth; in the ordered phase $c^{*}=1$, so $u_{1}^{*}=u_{2}^{*}$ and $\xi_{c}^{-1}=-\log\chi_{\perp}$, hence $\chi_{\perp}\to 1$ implies $\xi_{c}\to\infty$. Consequently, the edge-of-chaos is primarily about $\xi_{c}$ rather than $\xi_{q}$, except in special cases. For \relu{}, the curvature vanishes everywhere away from the origin, leading to a coincidence between $\xi_c$ and $\xi_q$, both diverging as $\chi_{\perp}\to 1$, provided a finite $q^{*}$ exists. This is one concrete reason why \relu{}-style near-critical initializations are comparatively forgiving in practice.

The finite-width regime of interest is \(L/N \ll 1\), where these corrections remain small \cite{roberts2022principlesdeeplearningtheory}. The correlation length diverges as we approach criticality, amplifying finite-width and other non-Gaussian corrections. Thus, when probing critical power laws with dropout, we keep the dropout rate small but non-negligible: infinitesimal dropout would require prohibitively deep networks to estimate the correlation length and related quantities.

The same Gaussian activation kernel appears in infinite-channel CNNs and residual branches, although their depth dynamics differ (App.~\ref{app:cnn_extension}). In ResNets, each residual block adds its input back to its output, changing how correlations propagate through depth \cite{yang2017meanfieldresnets}. Our Transformer experiments test whether the scheduling idea transfers beyond the MLP theory (Sec.~\ref{sec:experiments}).

In App.~\ref{app:experimental_info} we test different architectures, datasets, and schedules. We focus on overparameterized regimes where dropout improves generalization and accuracy, and thus where it is used in practice, and we explicitly study the trade-off between dropout-driven regularization and stable signal propagation. When we approximate propagation using the local fixed point at each layer, as below, concentrating a fixed budget into a step improves propagation regardless of where the step is placed. To choose its position, we use regularization reach: with the same decay scale for all mask positions, early masks expose more later layers to their perturbation. This motivates testing whether the first layers are the most useful place to inject noise.

\section{Universality Classes and Critical Scaling under Dropout}

The following analysis provides the mathematical basis for our scheduling argument, with results on critical scaling and universality that may be of particular interest to physicists. Readers primarily interested in experimental performance can skip directly to Sec.~\ref{sec:experiments}.

Dropout \cite{srivastava2014dropout,wager2013dropout} is often used as a regularizer that reduces overfitting by randomly masking activations during training, thereby reducing the network's reliance on any single subset of pathways and discouraging co-adaptation. In the previous literature, dropout was explored in MFT \cite{lee2019mean, schoenholz2017deepinformationpropagation}, but not in the universality class sense. We apply dropout after the activation as multiplicative Bernoulli noise,
\begin{equation}
p^{(\ell)}_{j} \sim \mathrm{Bernoulli}(\rho), \qquad y^{(\ell)}_j \;\mapsto\; \tilde y^{(\ell)}_j \equiv \frac{p^{(\ell)}_{j}}{\rho} y^{(\ell)}_j,
\end{equation}
i.e.\ the standard inverted-dropout convention in which $\mathbb E_p[\tilde y^{(\ell)}_j]=y^{(\ell)}_j$, and hence $\mathbb E_p[\tilde z^{\ell+1}_i\mid W,b,y^\ell]=z^{\ell+1}_i$, so the mean preactivation is invariant to the dropout rate \cite{schoenholz2017deepinformationpropagation}. Throughout, $\rho$ is the keep probability. When comparing two inputs, we draw the dropout masks independently for each input\footnote{If the two inputs share a mask, the same coordinates are removed, so identical representations remain identical and $c=1$ is preserved. Here we consider independently sampled masks.}; this independence is the source of the decorrelation effect analyzed below.

Tanh has rapidly decaying Hermite coefficients, whereas \relu{} has a power-law tail. The tails determine which derivatives of the correlation map remain finite at perfect alignment and thus the dropout scaling laws (App.~\ref{sec:var-opt-act}). This classification concerns regularity, rather than the scale invariance considered by \citet{roberts2022principlesdeeplearningtheory}.

Consider sending exactly the same input through the network twice. Without dropout, its representations remain identical at every layer, whereas independent masks remove different coordinates on the two passes and break this alignment. Rescaling the surviving activations preserves their mean over masks, but does not restore the agreement between the two representations. Perfect alignment therefore ceases to be a fixed point, cutting off the divergence of the depth correlation length at the edge \cite{schoenholz2017deepinformationpropagation}. For a single input, the corresponding variance increase can instead be absorbed into an effective weight variance, $\sigma_w^2\mapsto\sigma_w^2/\rho$.

With keep probability $\rho$, the preactivations become
\begin{equation}
z_i^\ell = \frac{1}{\rho} \sum_{j=1}^{N_{\ell-1}} W_{ij}^\ell p_j^{\ell-1} y_j^{\ell-1} + b_i^\ell.
\end{equation}
The single-input variance recursion becomes
\begin{equation}
\bar{q}_{aa}^{\ell} = \frac{\sigma_{w}^{2}}{\rho} \int Dz \phi^{2}\left( \sqrt{\bar{q}_{aa}^{\ell-1}} z \right) + \sigma_{b}^{2},
\end{equation}
where the bar denotes the variance with dropout. A fixed point, if it exists, satisfies
\begin{equation}
\bar q^{*} \;=\; \frac{\sigma_w^2}{\rho}\int Dz \phi^2\big(\sqrt{\bar q^{*}} z\big) + \sigma_b^2.
\label{eq:qstar-dropout}
\end{equation}
Assuming a positive fixed point $\bar q^*$, we can evaluate the correlation map at perfect alignment to find how far independent masks move the representations away from $c=1$. For two distinct inputs $a\neq b$ with $c^{\ell}_{ab}=1$, we obtain \cite{schoenholz2017deepinformationpropagation}
\begin{equation}
\bar F_\rho(1)= \bar c^{\ell+1}_{ab} = 1 - \frac{1 - \rho}{\rho \bar{q}^{*}} \sigma_{w}^{2} \int Dz \phi^{2}\left( \sqrt{\bar{q}^{*}} z \right).
\end{equation}
We call this departure from perfect alignment the dropout field
\begin{align}
h &\;\equiv\; 1-\bar F_\rho(1)\nonumber\\
&\;=\; \frac{1-\rho}{\rho \bar q^{*}}\; \sigma_w^2\int Dz \phi^2\big(\sqrt{\bar q^{*}} z\big).
\label{eq:Delta_exact}
\end{align}
For a nondegenerate channel with $\sigma_w^2\int Dz\,\phi(\sqrt{\bar q^*}z)^2>0$, construction gives $\rho<1\Rightarrow h>0$,
so the fixed point at $c^*=1$ is immediately spoiled: even if two inputs are perfectly aligned at some depth, the next layer produces $\bar c_{ab}=1-h<1$. For weak dropout, writing $\delta\rho\equiv 1-\rho$ with $\delta\rho\ll 1$, suppose the no-dropout variance fixed point is positive, isolated and stable, with variance-map derivative $|\chi_q|<1$. The implicit-function theorem then gives $\bar q^*=q^*+\mathcal O(\delta\rho)$, so to leading order
\begin{equation}
h = \delta\rho\; \frac{\sigma_w^2}{\bar q^{*}}\int Dz \phi^2\big(\sqrt{\bar q^{*}} z\big) + \mathcal O(\delta\rho^2).
\label{eq:Delta_weakdrop}
\end{equation}
This variance expansion excludes a marginal variance fixed point. For bias-free \relu{} at unchanged He initialization, dropout gives $q_{\ell+1}=q_\ell/\rho$; its normalized correlation map can still be analyzed without a converged variance, as in App.~\ref{sec:kinks}.
The backward pass has the same source of asymmetry. Let \(\delta^\ell_{j,a}\) be the backpropagated preactivation gradient for input \(a\). With inverted dropout,
\begin{equation}
\delta^\ell_{j,a} = \frac{p^\ell_{j,a}}{\rho} \phi'(z^\ell_{j,a}) \sum_i W^{\ell+1}_{ij}\delta^{\ell+1}_{i,a}.
\end{equation}
For equal hidden-layer widths, let \(G^\ell_{ab}\equiv \mathbb E[\delta^\ell_{j,a}\delta^\ell_{j,b}]\) denote the covariance per coordinate. Under the usual gradient-independence approximation at infinite width, the recursion contains
\begin{equation}
G^\ell_{ab} = \sigma_w^2 \mathbb E\left[ \frac{p^\ell_a p^\ell_b}{\rho^2} \phi'(u_a)\phi'(u_b) \right] G^{\ell+1}_{ab}.
\end{equation}
Since $p_a^2=p_a$, the mask contributes
\[
\frac{\mathbb E[p_a^2]}{\rho^2}=\frac1\rho,
\qquad
\frac{\mathbb E[p_ap_b]}{\rho^2}=1\quad(a\ne b).
\]
Within the gradient-independence approximation, the variance therefore receives an extra factor that is absent from the covariance between different inputs. After normalization, this suggests a loss of gradient alignment, described by a backward field $h_\nabla(\rho)$. Its scaling requires a separate calculation: the backward kernel contains $\phi'$, which is discontinuous for \relu{}, so we cannot simply transfer the forward result.

\subsection{Activation Regularity and Critical Exponents}

With masks drawn independently across units and inputs, we measure the fixed point's departure from perfect correlation by $m\equiv1-c^*$, and its detuning from the correlation edge by $t\equiv\chi_\rho-1$, where $\chi_\rho\equiv\bar F_\rho'(1)$ is the slope at perfect alignment. The dropout field $h$ is defined in \eqref{eq:Delta_exact}. For smooth activations, Price's theorem gives
\begin{align}
\chi_\rho &= \sigma_w^2\int Dz \big[\phi'(\sqrt{\bar q^{*}} z)\big]^2,
\label{eq:chi_dropout}\\
g_\rho &\equiv \bar F_\rho''(1) = \sigma_w^2 \bar q^{*}\int Dz \big[\phi''(\sqrt{\bar q^{*}} z)\big]^2.
\label{eq:g_dropout}
\end{align}
We assume the derivative identities hold through third order, with $\phi',\phi'',\phi^{\prime\prime\prime}\in L^2(\mathcal N(0,\bar q^*))$, and that $\bar q^*$ and $g_\rho$ have finite positive limits near the reference edge (App.~\ref{app:price_dropout}). Expanding the fixed-point condition $1-m=\bar F_\rho(1-m)$ then gives, to leading order,
\begin{equation}
h = \frac{g_\rho}{2}m^2-tm.
\label{eq:EoS_correct}
\end{equation}
The omitted term is $\mathcal O(m^3)$; when $g_\rho=0$, as for a linear activation, this smooth-class balance does not apply.

For \relu{}, imagine two nearly identical signals arriving on opposite sides of zero: one passes through, while the other is clipped. At a fixed nonzero Gaussian variance and correlation $c$ close to one, the signals have a root-mean-square separation $\mathcal O(\sqrt{1-c})$, so a fraction $\mathcal O(\sqrt{1-c})$ can straddle the kink. Within this small region, their mean squared activation difference is $\mathcal O(1-c)$, giving a contribution $\mathcal O((1-c)^{3/2})$. A continuous, piecewise $C^2$ activation with a finite slope jump and locally bounded, Gaussian-integrable regular second derivative therefore has the leading balance
\begin{equation}
h=\kappa m^{3/2}-tm+\mathcal O(m^2),
\label{eq:eos_kink_again}
\end{equation}
where $\kappa>0$ is the kink coefficient; for the normalized bias-free \relu{} map at the zero-dropout edge, $\kappa=2\sqrt{2}/(3\pi)$ (App.~\ref{app:tube_argument_kink}). At $t=0$, the two balances give $m\sim h^{1/2}$ for smooth activations and $m\sim h^{2/3}$ for kinked activations.

Let $\lambda\equiv\bar F_\rho'(c^*)$ be the slope at the displaced fixed point. For $0<\lambda<1$, small correlation perturbations decay over the depth scale $\xi_{c,\rho}\equiv-1/\log\lambda$. At $t=0$, linearizing the two maps gives
\begin{equation}
\xi_{c,\rho}(0,h)\sim
\begin{cases}
h^{-1/2}, & \text{smooth activations},\\
h^{-1/3}, & \text{kinked activations}.
\end{cases}
\end{equation}
Independent dropout therefore makes the propagation length finite at the edge, and whether the activation is smooth or kinked determines how this length depends on dropout. It measures how quickly the network loses sensitivity to input correlations at initialization, which we use below to compare dropout schedules.

Table~\ref{tab:critical_exponents} collects our exponents: $\xi(t,0)\sim|t|^{-\nu_t}$, $\xi(0,h)\sim h^{-\nu_\rho}$, $m(t,0)\sim t^\beta$ for $t>0$, $m(0,h)\sim h^{1/\delta}$, and $\left.\partial_hm\right|_{h\to0}\sim|t|^{-\gamma}$, with $\xi$ denoting $\xi_{c,\rho}$. At $t=h=0$, $m_\ell\equiv1-c_\ell\sim\ell^{-\theta_{\rm rel}}$ defines the relaxation exponent. The exponent $\alpha$ describes $f_{\rm sing,on}(t,0)\sim t^{2-\alpha}$ as $t\to0^+$, where $f_{\rm sing,on}$ is the singular stationary value of a potential obtained by integrating the equation of state. App.~\ref{ap:exponent_derivations} gives the definitions and derivations.

\begin{table}[h]
\caption{Critical exponents for the two activation classes at initialization. The $\nu$ exponents measure depth correlation length. Ordinary bias-free \relu{} dropout follows $t=-h$, retaining the leading $\xi\sim h^{-1/3}$ law of the kinked field direction (App.~\ref{ap:exponent_derivations}).}
\label{tab:critical_exponents}
\centering
\small
\setlength{\tabcolsep}{3pt}
\renewcommand{\arraystretch}{1.25}
\begin{tabular}{@{} l ccccccc @{}}
\toprule
\textbf{Universality Class} & $\boldsymbol{\nu_t}$ & $\boldsymbol{\beta}$ & $\boldsymbol{\theta}_{\rm rel}$ & $\boldsymbol{\gamma}$ & $\boldsymbol{\delta}$ & $\boldsymbol{\nu_\rho}$ & $\boldsymbol{\alpha}$ \\
\midrule
\textbf{Smooth activation} & $1$ & $1$ & $1$ & $1$ & $2$ & $\frac{1}{2}$ & $-1$ \\[6pt]
\textbf{Kinked activation} & $1$ & $2$ & $2$ & $1$ & $\frac{3}{2}$ & $\frac{1}{3}$ & $-3$ \\
\bottomrule
\end{tabular}
\end{table}

\begin{figure*}[ht]
 \centering
 \includegraphics[width=1.0 \textwidth]{\PaperRoot/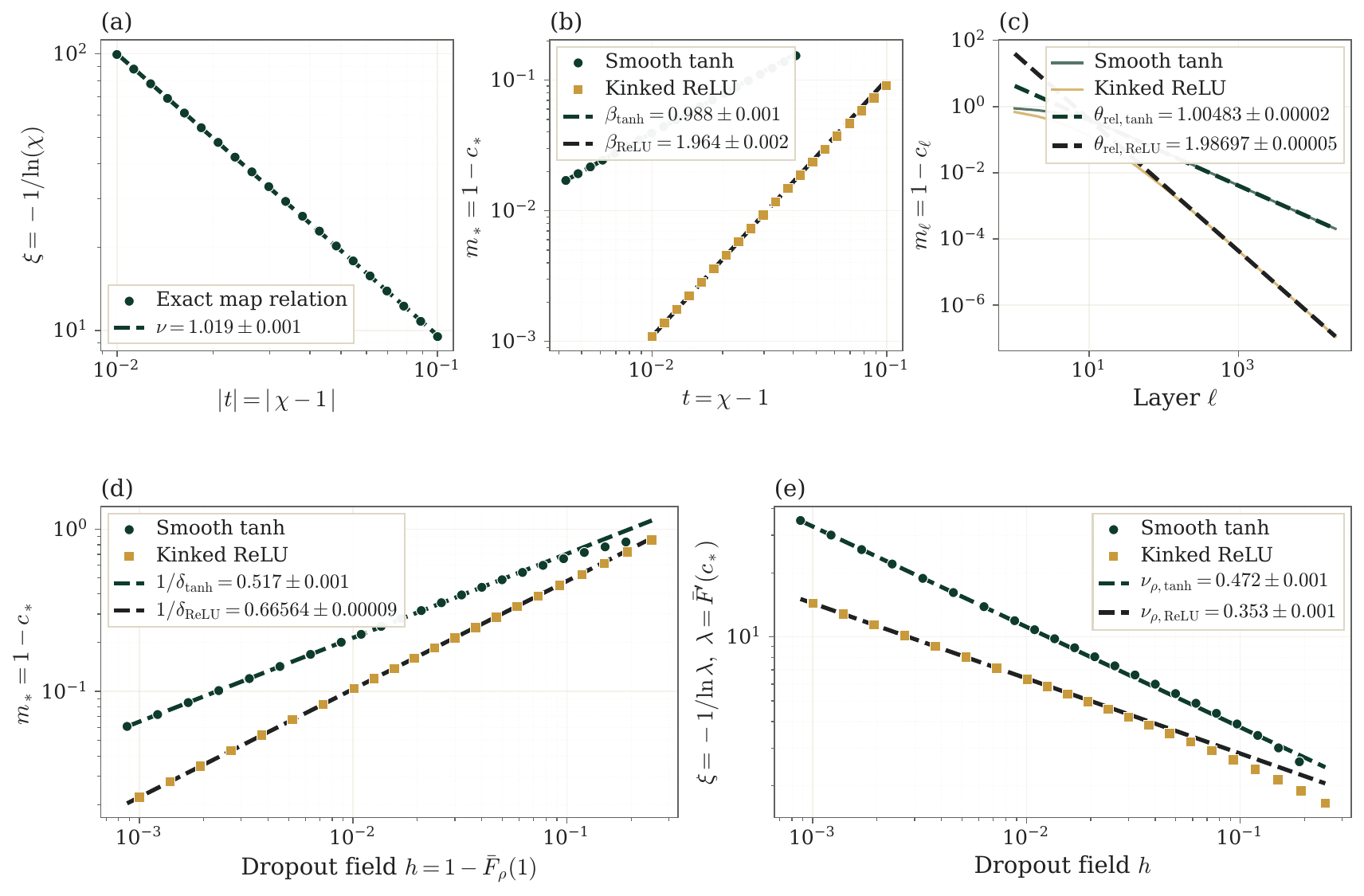}
\caption{
Critical scaling from the mean-field calculation. The top row shows the depth scale $-1/\log\chi$ on the ordered side, fixed-point decorrelation above the edge, and relaxation at criticality. The bottom row varies the dropout field at $\chi=1$. Dashed lines are log--log fits; quoted errors are regression standard errors on deterministic curves. Finite fitting windows and numerical approximation errors are not included. The tunable \relu{} channel is the formal family specified in App.~\ref{app:experimental_info}, rather than an independently tunable finite-variance \relu{} initialization.}
 \label{fig:dropout_scaling}
\end{figure*}

Figure~\ref{fig:dropout_scaling} checks these power laws by iterating the mean-field map and shows where the curves depart from them as we move away from the asymptotic regime.

\subsection{Scaling collapse}
\begin{figure*}[ht]
 \centering
 \includegraphics[width=1.0\textwidth]{\PaperRoot/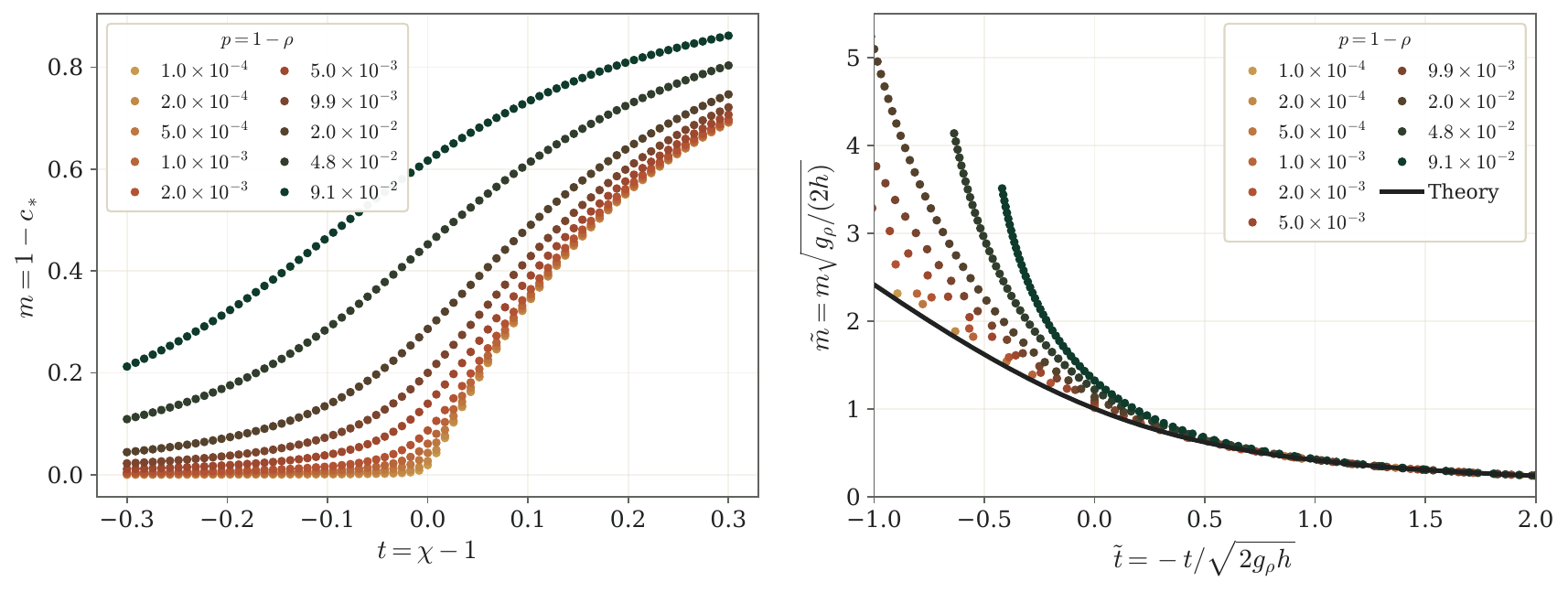}
 \caption{
 Two-parameter crossover and scaling collapse for the smooth universality class (tanh), computed from the full mean-field map. Each curve fixes the dropout probability $p=1-\rho$; its field $h$ and curvature $g_\rho$ are recomputed at each point. After rescaling, near-critical points approach the leading equation-of-state curve, while larger fields and decorrelations show departures from it. The kinked counterpart is shown in Fig.~\ref{fig:kinked_collapse}.}
 \label{fig:uni_scaling}
\end{figure*}

Changing dropout changes both the decorrelation and the distance from criticality at which dropout becomes important. The equation of state tells us how to rescale these quantities so that curves obtained at different dropout strengths coincide near criticality. For smooth activations, we define
\begin{equation}
\tilde m \equiv m \sqrt{\frac{g_\rho}{2h}}, \qquad \tilde t \equiv -\frac{t}{\sqrt{2g_\rho h}}.
\label{eq:rescaled_vars_smooth}
\end{equation}
giving
\begin{equation}
\tilde m = \sqrt{1+\tilde t^{2}}-\tilde t.
\label{eq:universal_scaling_function_smooth_correct}
\end{equation}
The leading equation therefore gives the same curve at every dropout strength \cite{ZinnJustin2002QFTCritical}, which the full correlation map approaches near criticality (Fig.~\ref{fig:uni_scaling}). The crossover occurs at $|t|\sim\sqrt{g_\rho h}$ for smooth activations and $|t|\sim\kappa^{2/3}h^{1/3}$ for kinked activations, separating the region where dropout sets the decorrelation from the region where distance to criticality matters more. App.~\ref{app:crossover_details} gives both scaling functions and their limiting branches.

For ordinary bias-free \relu{}, the microscopic dropout path locks $t=-h$, so dropout controls the leading response as $h\to0$, without allowing us to vary the two parameters independently along $t=0,h>0$. Higher-order terms in the map and changes in the kink coefficient cause the curves to depart from the leading prediction at larger fields. These crossover scales indicate how close an initialization must be to the edge before dropout controls its propagation length; they do not by themselves predict trained performance.

\subsection{Optimal Dropout Scheduling and Regularization Reach}
\label{subsec:optimal_dropout}

Thus far, we have considered only homogeneous dropout schedules, where dropout is treated as a constant hyperparameter in depth. We now seek profiles that will keep or improve the regularization, while aiding information propagation through longer correlation lengths. Let us introduce a depth-dependent keep probability $\rho_\ell$ for the mask applied after the activation at layer $\ell$ and define the associated dropout field
\begin{equation}
h_\ell \;\equiv\; 1-\bar F_{\rho_\ell}(1),
\end{equation}
i.e. the shift of the correlation map at perfect alignment induced by keep probability $\rho_\ell$. If variances have settled so that a scalar map applies at each layer, the exact linear response along the trajectory $c_{\ell+1}=\bar F_{\rho_\ell}(c_\ell)$ is
\begin{equation}
\delta c_{L+1}=\left[\prod_{\ell=1}^L \bar F_{\rho_\ell}'(c_\ell)\right]\delta c_1.
\end{equation}
To compare schedules, we approximate each derivative by its value at the local fixed point and hold the near-critical coefficients $t$ and $g_\rho$ at the same reference values. In the smooth class it follows that (see \eqref{eq:lambda_smooth_correct})
\begin{equation}
\lambda(t,h_\ell)\;\simeq\; 1-\sqrt{t^2+2g_\rho h_\ell}.
\end{equation}
For $0<\lambda_\ell\lesssim 1$, this approximates the propagated perturbation by
\begin{equation}
\delta c_{L+1} \;\sim\; \prod_{\ell=1}^L \lambda_\ell \delta c_1, \qquad \lambda_\ell \;\equiv\; \lambda(t,h_\ell),
\label{delta_c_dropout_perturb}
\end{equation}
so we define the effective correlation length as
\begin{equation}
\xi_{\rm eff}^{-1} \;\equiv\; -\frac{1}{L}\sum_{\ell=1}^L \log\lambda_\ell \;\simeq\; \frac{1}{L}\sum_{\ell=1}^L \sqrt{t^2+2g_\rho h_\ell}.
\end{equation}
At the correlation edge $t=0$, this reduces to
\begin{equation}
\xi_{\rm eff}^{-1}(t=0) \;\simeq\; \sqrt{2g_\rho} \frac{1}{L}\sum_{\ell=1}^L h_\ell^{1/2} \;\propto\; \frac{1}{L}\sum_{\ell=1}^L h_\ell^{1/2}.
\label{eff_corr}
\end{equation}
Fix $\bar h=L^{-1}\sum_\ell h_\ell$ and $0\le h_\ell\le h_{\max}$, with $0<\bar h\le h_{\max}$. Jensen's inequality and the chord bound give
\begin{equation}
\frac{\bar h}{\sqrt{h_{\max}}}
\le \frac1L\sum_{\ell=1}^L\sqrt{h_\ell}
\le \sqrt{\bar h}.
\end{equation}
Uniform dropout therefore maximizes the leading-order decay rate and minimizes $\xi_{\rm eff}$. In the continuum, the minimum decay rate uses only $0$ and $h_{\max}$, with active fraction $f=\bar h/h_{\max}$. Its position is unrestricted; regularization reach below selects the input side:
\begin{equation}
h(x)=\begin{cases}h_{\max},&0\le x\le f,\\
0,&f<x\le1,\end{cases}
\qquad x=\ell/L.
\end{equation}
For finite $L$, at most one layer remains partially filled. The field budget agrees with a dropout-probability budget to leading order in weak dropout, by \eqref{eq:Delta_weakdrop}. These conclusions assume weak fields and local fixed points; abrupt schedules need not preserve this approximation.

Comparing this step profile with uniform dropout in the continuum, to leading order uniform dropout yields $\xi_{\rm eff}\sim \bar h^{-1/2}$ while the step schedule gives
\begin{equation}
\frac{\xi_{\rm eff,step}}{\xi_{\rm eff,uniform}} = \sqrt{\frac{h_{\max}}{\bar h}} \;\ge\; 1.
\end{equation}
For kinked activations at $t=0$, the same local approximation gives $\xi_{\rm eff}^{-1}\propto L^{-1}\sum_\ell h_\ell^{1/3}$. Jensen's inequality and the chord bound now give
\begin{equation*}
\frac{\bar h}{h_{\max}^{2/3}}
\le \frac1L\sum_{\ell=1}^L h_\ell^{1/3}
\le \bar h^{1/3}.
\end{equation*}
Concentrating dropout therefore again maximizes the approximate $\xi_{\rm eff}$, with at most one partially filled layer; regularization reach selects the earliest layers.

\begin{figure*}[t]
\centering
\includegraphics[width=0.93\textwidth]{\PaperRoot/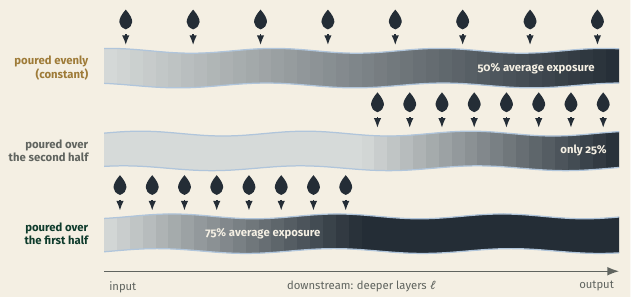}
\caption{Imagine you have a flowing river and a set amount of ink, with the goal of causing as much corruption as possible at every point: where would you drop it?}
\label{fig:river_reach}
\end{figure*}

Dropout is used to ``corrupt'' the network so as to prevent it from memorizing certain ``paths'' from input to output. Imagine a river where a fixed amount of ink is to be distributed to corrupt every cross section as much as possible. Reframed this way, front-loading lets the current carry the ink through the rest of the river. Regularization reach makes this intuition explicit.

Let us normalize depth $x=\ell/L$ and write $j(x)=h(x)/\bar h$ for the injection density, with $\bar h>0$ and $\int_0^1j(x)\,dx=1$. For the river in Fig.~\ref{fig:river_reach}, where the ink does not fade as it travels downstream, the accumulated exposure at $x$ and its average along the river are
\begin{align}
E(x)&=\int_0^x j(s)\,ds,\\
\bar E&=\int_0^1 E(x)\,dx
       =\int_0^1(1-s)j(s)\,ds.
\end{align}
Uniform injection has $j=1$ throughout; early and late injection have $j=2$ over their respective halves. Thus
\begin{equation}
\bar E_{\rm early}=\tfrac34,\qquad
\bar E_{\rm uniform}=\tfrac12,\qquad
\bar E_{\rm late}=\tfrac14.
\end{equation}
All three schedules deliver the same total ink to the outlet, but expose different amounts of the river along the way.

An early mask affects many later layers, even as its influence fades; a late mask has fewer layers left to regularize. Compare the same signal with and without a mask at layer $\ell$, writing $c_k$ for their normalized second-moment overlap at layer $k$ and $\eta_k=1-c_k$. Inverted dropout preserves the cross moment but multiplies the masked second moment by $1/\rho_\ell$, so
\begin{equation}
\eta_\ell=1-\sqrt{\rho_\ell}
=\frac{1-\rho_\ell}{2}+\mathcal O((1-\rho_\ell)^2).
\end{equation}
By \eqref{eq:Delta_weakdrop}, this is $\eta_\ell\simeq C h_\ell$ for a constant $C>0$. Neglecting variance transients, linearizing the ordered correlation map with $\lambda=F'(1)\in(0,1)$ gives
\begin{align}
\eta_{k+1}&\simeq\lambda\eta_k,\\
\eta_{\ell+r}&\simeq\lambda^r\eta_\ell
\simeq C h_\ell e^{-r/\xi_c},\qquad
\xi_c=-\frac{1}{\log\lambda}.
\end{align}
Holding the same background, $C$ and $\xi_c$ fixed across mask positions, we add this influence over the remaining layers to define the exposure:
\begin{align}
\mathcal D_\ell\equiv\sum_{r=1}^{L-\ell}\eta_{\ell+r}
&\simeq C h_\ell\sum_{r=1}^{L-\ell}e^{-r/\xi_c}\nonumber\\
&\approx C h_\ell\xi_c\left(1-e^{-(L-\ell)/\xi_c}\right)\nonumber\\
&\equiv C h_\ell w_\ell.
\label{eq:regularization_reach}
\end{align}
The second line replaces the sum by its continuum integral. Since $w_\ell$ decreases with depth, moving dropout to earlier layers increases the summed exposure $\sum_\ell h_\ell w_\ell$. At fixed budget and per-layer bounds, its maximum fills the earliest layers first. Adding mask effects in this way is our regularization heuristic; whether the resulting schedule improves generalization is tested experimentally. App.~\ref{app:regularization_reach} gives further details and the limits of the approximation.

\section{Experiments}
\label{sec:experiments}

\begin{table*}[t]
\centering
\caption{Improvement over uniform dropout on CIFAR, keeping mean dropout at $\bar p=0.1$. Positive values mean lower test loss or higher test accuracy. Loss is cross-entropy; brackets give nominal paired $95\%$ confidence intervals.}
\label{tab:loss_improvements}
\footnotesize
\setlength{\tabcolsep}{4pt}
\renewcommand{\arraystretch}{1.15}
\begin{tabular}{@{}lrrlrrr@{}}
\toprule
& & & & Lowest test loss & \multicolumn{2}{c}{End of training} \\
\cmidrule(lr){5-5}\cmidrule(l){6-7}
\shortstack[l]{Dataset /\\model} & Epochs & Seeds & \shortstack[l]{Dropout\\placement} & \shortstack{Loss reduction\\(\%)} & \shortstack{Loss reduction\\(\%)} & \shortstack{Accuracy gain\\(percentage points)} \\
\midrule
\shortstack[l]{CIFAR-10\\MLP (ReLU)} & 75 & 25 & \shortstack[l]{First 3 layers\\$p=0.2$} & \shortstack{$+1.42$\\[-1pt]{\scriptsize $[1.19, 1.65]$}} & \shortstack{$+17.86$\\[-1pt]{\scriptsize $[17.31, 18.41]$}} & \shortstack{$+0.83$\\[-1pt]{\scriptsize $[0.55, 1.11]$}} \\
\addlinespace[4pt]
\shortstack[l]{CIFAR-10\\MLP (ReLU)} & 75 & 25 & \shortstack[l]{First 2 layers\\$p=0.3$} & \shortstack{$+2.07$\\[-1pt]{\scriptsize $[1.80, 2.33]$}} & \shortstack{$+22.61$\\[-1pt]{\scriptsize $[22.06, 23.15]$}} & \shortstack{$+1.08$\\[-1pt]{\scriptsize $[0.85, 1.31]$}} \\
\addlinespace[4pt]
\shortstack[l]{CIFAR-10\\MLP (ReLU)} & 125 & 10 & \shortstack[l]{First 2 layers\\$p=0.3$} & \shortstack{$+2.54$\\[-1pt]{\scriptsize $[2.03, 3.06]$}} & \shortstack{$+34.90$\\[-1pt]{\scriptsize $[34.36, 35.44]$}} & \shortstack{$+1.56$\\[-1pt]{\scriptsize $[1.22, 1.89]$}} \\
\addlinespace[4pt]
\shortstack[l]{CIFAR-10\\MLP (GELU)} & 75 & 10 & \shortstack[l]{First 2 layers\\$p=0.3$} & \shortstack{$+2.45$\\[-1pt]{\scriptsize $[1.90, 3.00]$}} & \shortstack{$+29.79$\\[-1pt]{\scriptsize $[28.96, 30.61]$}} & \shortstack{$+0.62$\\[-1pt]{\scriptsize $[0.16, 1.08]$}} \\
\addlinespace[4pt]
\shortstack[l]{CIFAR-100\\ViT} & 75 & 10 & \shortstack[l]{Linear decrease\\$0.2\to0$} & \shortstack{$+1.39$\\[-1pt]{\scriptsize $[0.56, 2.21]$}} & \shortstack{$+4.15$\\[-1pt]{\scriptsize $[3.40, 4.90]$}} & \shortstack{$+0.66$\\[-1pt]{\scriptsize $[0.36, 0.96]$}} \\
\addlinespace[4pt]
\shortstack[l]{CIFAR-10\\ViT} & 75 & 10 & \shortstack[l]{First 6 blocks\\$p=0.2$} & \shortstack{$+1.51$\\[-1pt]{\scriptsize $[0.33, 2.68]$}} & \shortstack{$+7.40$\\[-1pt]{\scriptsize $[5.35, 9.41]$}} & \shortstack{$+0.68$\\[-1pt]{\scriptsize $[0.24, 1.12]$}} \\
\bottomrule
\end{tabular}
\par\smallskip\begin{minipage}{\textwidth}\scriptsize Lowest test loss is the minimum reached by each run; final results use the epoch count shown. Loss reductions compare seed means, $100(L_U-L_F)/L_U$; accuracy gains are absolute differences. Profiles were chosen by mean final test loss and kept unchanged when adding seeds; the epochs with the lowest loss also use test data. Intervals use Fieller's method for loss and Student-$t$ for accuracy, without correction for selection.\\[3pt] All MLPs have six hidden layers of width 256. The first two rows share their uniform baseline. The 125-epoch ReLU comparison also changes initialization and batch size (App.~\ref{app:sweeps}). The CIFAR-10 ViT applies dropout to both attention and MLP branches.\end{minipage}
\end{table*}

Our propagation calculation favours concentrating dropout in fewer layers, while regularization reach favours placing those layers near the input, where more of the network is exposed to each mask. We test this by sliding a dropout block from front-loaded to back-loaded, redistributing the same average dropout on CIFAR, and tuning uniform and front-loaded networks independently on speech, image and financial data.

We compare uniform dropout with three front-loaded profiles. At mean dropout $\bar p=0.1$, \emph{step early} applies $p=0.2$ to the first half of the layers, \emph{big step} applies $p=0.3$ to the first third, and \emph{linear decreasing} de-ramps from $0.2$ to $0$. Apps.~\ref{app:experimental_info} and~\ref{app:benchmark_protocols} specify the profiles at other dropout budgets. On CIFAR we hold $\bar p$ fixed, so only its placement changes; profiles were chosen by mean final test loss and kept unchanged when adding seeds, while the epoch with the lowest loss is also chosen using test data. On Speech Commands \cite{warden2018speechcommands}, Tiny ImageNet \cite{le2015tinyimagenet} and FI-2010 \cite{ntakaris2018benchmark}, validation loss determines the dropout profile, learning rate, mean dropout and epoch used for test evaluation. Runs are paired by seed. The CIFAR tables and the comparisons at the best validation epoch use at least ten pairs; final evaluations on Speech Commands, Tiny ImageNet and FI-2010 use five or ten, while the block-placement and correlation experiments use five. We report nominal $95\%$ Fieller intervals for relative loss reductions \cite{fieller1954interval} and Student-$t$ intervals for accuracy differences, conditional on the split and settings and without correction for selection (App.~\ref{app:benchmark_protocols}).

\begin{figure*}[tp]
 \centering
 \includegraphics[width=0.95\textwidth]{\PaperRoot/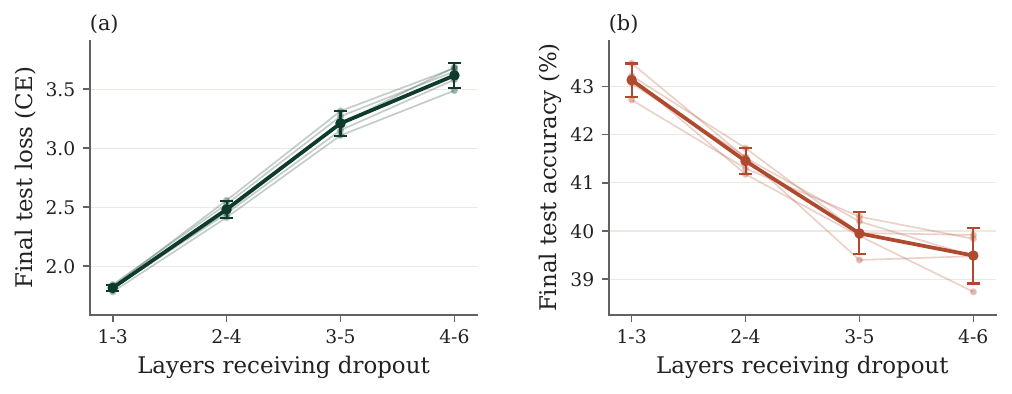}
 \caption{The same dropout block, moved from the input towards the output of a six-layer CIFAR-10 MLP. Three consecutive layers use $p=0.2$, with zero elsewhere, keeping average dropout at $0.1$. Faint lines follow the five seeds; thick lines and bars give means and pointwise $95\%$ Student-$t$ confidence intervals at epoch 75. App.~\ref{app:mlps} gives the protocol.}
 \label{fig:sliding_block}
\end{figure*}

\paragraph{Dropout block placement}
We show a monotonic effect of dropout-block placement by sliding the same block from front-loaded to back-loaded, with mean final test loss and accuracy improving monotonically as we approach our proposed front-loaded placement. We slide a three-layer block of $p=0.2$ dropout through a six-layer CIFAR-10 MLP, keeping the network, learning rate and $\bar p=0.1$ fixed (Fig.~\ref{fig:sliding_block}). In all five seeds, final test loss rises with every move towards the output; the mean rises from $1.82$ with the block at layers 1--3 to $3.62$ at layers 4--6. Over the same range, mean accuracy falls from $43.1\%$ to $39.5\%$.

\begin{figure*}[tp]
 \centering
 \includegraphics[width=0.95\textwidth]{\PaperRoot/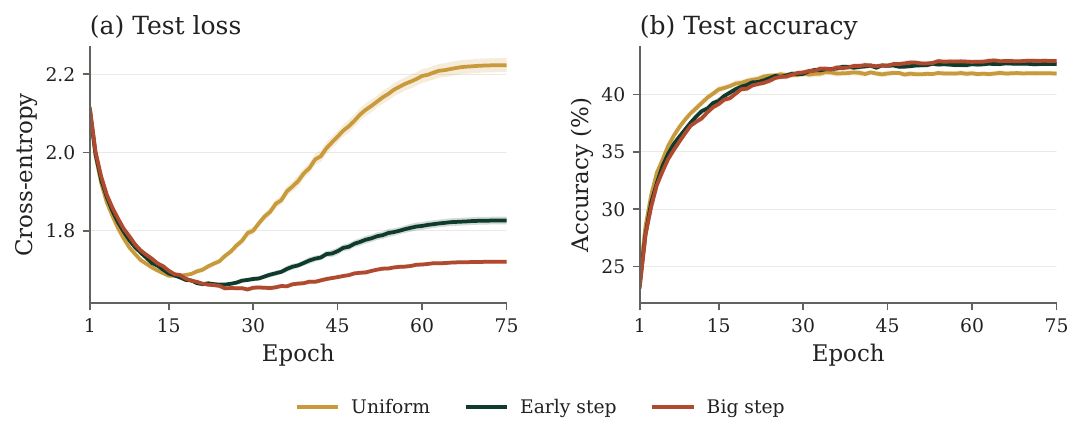}
 \caption{Test loss and accuracy during 75 epochs of CIFAR-10 training with a six-layer, width-256 ReLU MLP. Uniform dropout uses $p=0.1$ throughout; early step uses $p=0.2$ in the first three layers, and big step uses $p=0.3$ in the first two, with zero elsewhere. All three keep average dropout at $0.1$. Lines give means over 25 seeds and shading gives pointwise $95\%$ Student-$t$ confidence intervals. App.~\ref{app:mlps} gives the protocol and Fig.~\ref{fig:triple_dropout} includes the training curves and additional controls.}
 \label{fig:cifar_test_curves}
\end{figure*}

\begin{figure*}[tp]
 \centering
 \includegraphics[width=\textwidth]{\PaperRoot/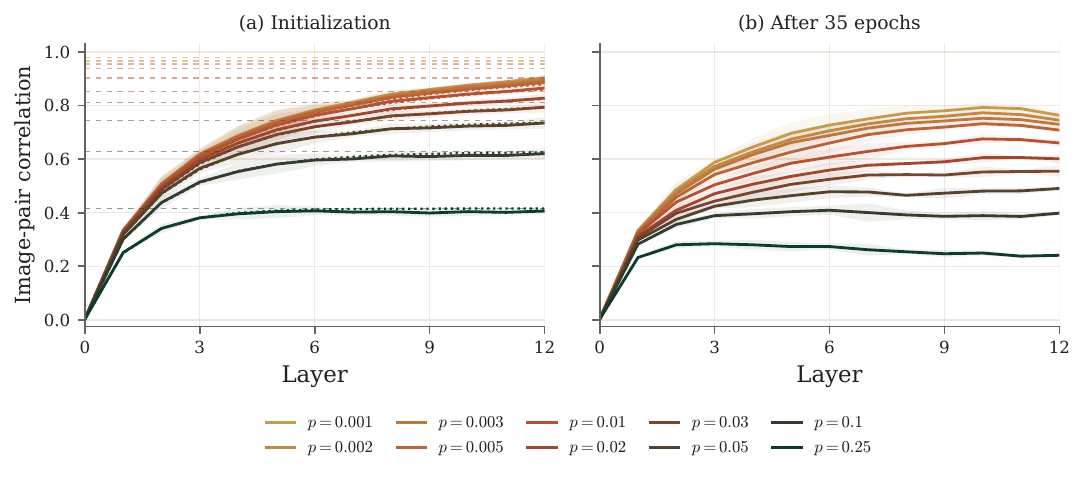}
 \caption{Correlations between distinct CIFAR-10 images before and after 35 training epochs in a twelve-layer \relu{} MLP with uniform dropout. Solid curves give five-seed means and shading gives pointwise $95\%$ confidence intervals. In (a), dotted curves give the initialization recursion and dashed lines its infinite-depth fixed points; at small dropout, the longer correlation length means that twelve layers are not enough to reach these values. Dropout is active in both panels. App.~\ref{app:cifar_correlations} gives the definition and protocol.}
 \label{fig:cifar_correlations}
\end{figure*}

\begin{table*}[t]
\centering
\caption{Improvements over uniform dropout, calculated from seed means and using the same profile across each row. Positive values mean lower test loss or higher test accuracy. Best means the epoch with the lowest validation loss in each run; final means the last epoch. Loss is cross-entropy; its reduction is $100(L_U-L_F)/L_U$, and accuracy gains are in percentage points (pp). Brackets give nominal paired 95\% confidence intervals, with the number of seed pairs, $n$, given for each evaluation. Profiles are chosen by their mean minimum validation loss among completed comparisons. Speech Commands holds out clips, with overlapping speakers. Learning rate and mean dropout are tuned separately for each profile, so average dropout can differ. App.~\ref{app:benchmark_methods} gives the selection procedure, available measurements and interval calculations.}
\label{tab:frontloaded_results}
\footnotesize
\setlength{\tabcolsep}{4pt}
\renewcommand{\arraystretch}{1.15}
\begin{tabular}{@{}llrrrrrr@{}}
\toprule
& & \multicolumn{3}{c}{Test at best validation epoch} & \multicolumn{3}{c}{Test at final epoch} \\
\cmidrule(lr){3-5}\cmidrule(l){6-8}
Experiment (training $N$) & Profile & $n$ & \shortstack{Loss reduction\\(\%)} & \shortstack{Accuracy gain\\(pp)} & $n$ & \shortstack{Loss reduction\\(\%)} & \shortstack{Accuracy gain\\(pp)} \\
\midrule
\shortstack[l]{Speech Commands (20,020)\\MLP} & Step early & 10 & \shortstack{$+12.56$\\[-1pt]{\scriptsize $[11.36, 13.73]$}} & \shortstack{$+1.72$\\[-1pt]{\scriptsize $[1.28, 2.16]$}} & 5 & \shortstack{$+12.59$\\[-1pt]{\scriptsize $[9.95, 15.13]$}} & \shortstack{$+1.71$\\[-1pt]{\scriptsize $[0.78, 2.64]$}} \\
\shortstack[l]{Speech Commands (20,020)\\Transformer} & Big step & 10 & \shortstack{$+9.04$\\[-1pt]{\scriptsize $[2.52, 15.24]$}} & \shortstack{$+2.06$\\[-1pt]{\scriptsize $[0.72, 3.41]$}} & 5 & \shortstack{$+15.51$\\[-1pt]{\scriptsize $[6.44, 23.04]$}} & \shortstack{$+1.10$\\[-1pt]{\scriptsize $[-0.10, 2.30]$}} \\
\addlinespace[3pt]
\shortstack[l]{Tiny ImageNet (20,000)\\MLP} & Step early & 10 & \shortstack{$+2.04$\\[-1pt]{\scriptsize $[1.72, 2.37]$}} & \shortstack{$+1.28$\\[-1pt]{\scriptsize $[0.99, 1.57]$}} & 5 & \shortstack{$+21.48$\\[-1pt]{\scriptsize $[19.59, 23.35]$}} & \shortstack{$+0.80$\\[-1pt]{\scriptsize $[0.47, 1.14]$}} \\
\shortstack[l]{Tiny ImageNet (20,000)\\Transformer} & Step early & 10 & \shortstack{$+2.21$\\[-1pt]{\scriptsize $[1.69, 2.73]$}} & \shortstack{$+1.49$\\[-1pt]{\scriptsize $[1.06, 1.92]$}} & 5 & \shortstack{$+21.06$\\[-1pt]{\scriptsize $[20.27, 21.83]$}} & \shortstack{$+0.73$\\[-1pt]{\scriptsize $[-0.19, 1.66]$}} \\
\shortstack[l]{Tiny ImageNet (80,000)\\MLP} & Step early & 10 & \shortstack{$+1.17$\\[-1pt]{\scriptsize $[0.92, 1.42]$}} & \shortstack{$+0.77$\\[-1pt]{\scriptsize $[0.50, 1.05]$}} & 10 & \shortstack{$+8.65$\\[-1pt]{\scriptsize $[8.26, 9.04]$}} & \shortstack{$+0.68$\\[-1pt]{\scriptsize $[0.48, 0.88]$}} \\
\shortstack[l]{Tiny ImageNet (80,000)\\Transformer} & Linear decreasing & 10 & \shortstack{$+0.76$\\[-1pt]{\scriptsize $[-0.24, 1.76]$}} & \shortstack{$+0.13$\\[-1pt]{\scriptsize $[-0.41, 0.67]$}} & 10 & \shortstack{$+1.66$\\[-1pt]{\scriptsize $[0.51, 2.81]$}} & \shortstack{$-0.00$\\[-1pt]{\scriptsize $[-0.41, 0.40]$}} \\
\addlinespace[3pt]
\shortstack[l]{FI-2010 (40,000)\\MLP} & Big step & 10 & \shortstack{$+4.25$\\[-1pt]{\scriptsize $[2.57, 5.88]$}} & \shortstack{$+0.81$\\[-1pt]{\scriptsize $[0.38, 1.25]$}} & 5 & \shortstack{$+68.23$\\[-1pt]{\scriptsize $[67.30, 69.12]$}} & \shortstack{$-2.48$\\[-1pt]{\scriptsize $[-2.86, -2.09]$}} \\
\bottomrule
\end{tabular}
\end{table*}

\paragraph{CIFAR at fixed average dropout.}
Redistributing the same budget towards the input lowers final test loss by $18$--$35\%$ for MLPs and $4$--$7\%$ for Vision Transformers \cite{dosovitskiy2021vit}, and raises final accuracy by $0.6$--$1.6$ percentage points (Table~\ref{tab:loss_improvements}). At the epoch where each run reaches its lowest test loss, the gap shrinks to $1.4$--$2.5\%$, so much of the benefit shows up late in training. GELU also benefits, with final test loss falling by $30\%$ versus $35\%$ for \relu{} at $\bar p=0.1$. In the \relu{} width sweep, both early schedules improve mean best test accuracy at widths 128--1024, where the large-$N$ approximation is better justified, while their differences from uniform are small at width 64. In the GELU sweep, the ordering by best test accuracy reverses at $\bar p=0.15$, where the local rates reach $0.30$ and $0.45$ (Figs.~\ref{fig:mlp_sweeps} and~\ref{fig:gelu_h_sweep}; App.~\ref{app:sweeps}). The theory covers MLPs only, yet front-loading also helps the tested Transformers. On CIFAR-100, linearly decreasing dropout raises final ViT accuracy from $48.6\%$ to $49.3\%$ (Fig.~\ref{fig:dropout_regimes_transformer}). On CIFAR-10, an early step in the attention block, the MLP block, or both gives higher mean final accuracy than uniform in each case (Fig.~\ref{fig:ablations}).

\paragraph{Speech Commands, Tiny ImageNet and FI-2010.}
\label{sec:extended_benchmarks}
On these datasets we use twelve-layer, width-256 models (App.~\ref{app:extended_experiments}). Table~\ref{tab:frontloaded_results} reports test loss and accuracy at the epoch with the lowest validation loss and at the final epoch. At the epoch with the lowest validation loss, Speech Commands gives the largest reductions in test loss, on a random clip split in which speakers overlap between training and test (App.~\ref{app:benchmark_methods}). In the five-seed MLP comparison, uniform dropout already lowers test loss by $11\%$ relative to no dropout. Across ten seed pairs, front-loading lowers test loss relative to uniform by $12.6\%$ for the MLP and $9.0\%$ for the Transformer, with accuracy gains of $1.7$ and $2.1$ percentage points. On Tiny ImageNet, test loss falls by around $2\%$ with 20,000 training examples and $1\%$ with 80,000. The ViT trained on 80,000 images is the only configuration in this table whose interval for the reduction in test loss includes zero. On FI-2010, the MLP's test loss improves by $4.3\%$ and its accuracy by $0.8$ percentage points.

\paragraph{Loss and accuracy late in training.}
The largest differences appear after the loss minimum. In the 25-seed CIFAR-10 budget comparison, uniform dropout's mean test loss rises $32\%$ from its minimum to the final epoch, against $4\%$ for big step (Fig.~\ref{fig:cifar_test_curves}). Meanwhile, big step's mean accuracy keeps improving, from $41.9\%$ at epoch 29 to $42.9\%$ at epoch 75. At the final epoch, the reduction in test loss also reaches $21\%$ on Tiny ImageNet with 20,000 training examples (Table~\ref{tab:frontloaded_results}). Loss and accuracy can also disagree: on FI-2010, front-loading cuts final loss by $68\%$ but lowers accuracy by $2.5$ percentage points.

\paragraph{Correlations before and after training.}
The theory describes networks at initialization. We therefore measure correlations between distinct CIFAR-10 images in a twelve-layer MLP before and after training, with dropout active in both measurements (Fig.~\ref{fig:cifar_correlations}). At initialization, the curves follow the mean-field recursion. At small $p$, the longer correlation length means that twelve layers are not enough to reach the predicted fixed point. With the dropout rate held at $p=0.001$, training lowers the mean correlation at layer 12 from $0.90$ to $0.76$ after 35 epochs, moving it further from the initialization prediction of $0.98$.

\section{Conclusions}
Smooth and kinked activations respond differently to dropout, giving different scaling laws for how far input correlations propagate at initialization. At fixed average dropout field, our propagation approximation favours concentrating dropout in fewer layers, while regularization reach favours placing those layers near the input, where each mask can affect more of the network. We show front-loading can improve the model chosen by validation loss and restrain the rise in test loss when training continues. In the CIFAR-10 budget comparison, uniform dropout's mean test loss rises by $32.0\%$ from its minimum to the final epoch, compared with $4.3\%$ for big-step dropout, while the latter's accuracy continues to improve. Speech Commands shows clear gains in both architectures on held-out clips; whether these gains extend to unseen speakers remains to be tested. Across the broader benchmarks, MLP gains are more consistent, while the difference in test loss for the larger Tiny ImageNet Transformer remains inconclusive when we choose the models by validation loss. It remains to extend this mean-field analysis to attention mechanisms and include finite-width corrections.

\section{Limitations}
Our theory describes initialization. Training changes the weights and the correlation curves, as Fig.~\ref{fig:cifar_correlations} shows, so the initialization calculation alone cannot predict the trained network's performance. We assume wide MLPs, with finite-width corrections that remain small as they accumulate with depth through $L/N$. In the scheduling calculation, we approximate each layer by its local fixed point, while regularization reach adds the effects of separate masks using the same decay scale. Strong dropout and interactions between masks can make these approximations fail.

The Transformer experiments test whether the same scheduling idea works beyond the MLPs described by our theory. Appendix~\ref{app:cnn_extension} discusses related Gaussian kernels in CNNs and residual branches; how dropout changes propagation through the full network remains to be derived. The observed loss curves measure overfitting, rather than memorization directly.

\noindent\emph{Acknowledgments.} The author thanks Riccardo Penco for support and discussions, and the anonymous ICML reviewers for feedback that sharpened the scheduling discussion and the treatment of the kinked-class scaling path. This work is partially supported by the U.S. Department of Energy under grant DE-SC0010118.

\section*{Reproducibility Statement}
\phantomsection\label{sec:reproducibility}
The repository contains training code, saved measurements and scripts for plotting and confidence intervals. Appendix~\ref{app:experimental_info} gives the CIFAR protocols; App.~\ref{app:benchmark_protocols} gives the protocols for the other datasets and identifies settings that could not be recovered. The missing CIFAR-100 configuration is documented alongside its reported training settings in Table~\ref{tab:vit_config}.

\section*{Impact Statement}
The main practical benefit is improved performance at fixed training cost, or potentially less training to reach a given level of performance. We do not anticipate additional ethical risks.

\bibliography{bibliography}
\bibliographystyle{icml2026}

\newpage
\onecolumn

\appendix

\section{Mean-Field Background and Dropout Recursions}

\subsection{Notation Guide}
\label{app:notation}

The same few letters carry several nearby meanings, so we collect the main conventions here.
\begin{table}[h]
\caption{Notation used across the main text and appendices.}
\label{tab:notation_guide}
\centering
\small
\begin{tabular}{@{}ll@{}}
\toprule
Symbol & Meaning \\
\midrule
\(\rho_\ell\) & keep probability in layer \(\ell\) \\
\(p_{\mathrm{drop},\ell}=1-\rho_\ell\) & raw dropout probability \\
\(p_j^{(\ell)}\) & Bernoulli dropout mask variable \\
\(m=1-c^\ast\) & decorrelation order parameter at the fixed point \\
\(m_\ell=1-c_\ell\) & depth-dependent decorrelation used in relaxation laws \\
\(\chi_\perp\), \(\chi_1\) & no-dropout angular susceptibility at \(c=1\) \\
\(\chi_\rho\) & dropout-deformed angular susceptibility \(\bar F_\rho'(1)\) \\
\(\chi_M\) & response susceptibility \(\partial m/\partial h\) \\
\(\chi\), \(\chi_q\) & generic Hermite/angle susceptibility and variance-channel susceptibility \\
\(\theta_{\rm rel}\) & relaxation exponent, \(m_\ell\sim \ell^{-\theta_{\rm rel}}\) \\
\bottomrule
\end{tabular}
\end{table}
\FloatBarrier

\subsection{Mean-Field Theory Primer}
\label{app:mft_primer}

Fermi quoted von Neumann to Dyson: ``With four parameters I can fit an elephant, and with five I can make him wiggle his trunk'' \cite{Dyson2004Fermi}. Modern machine learning pushes this worry to an extreme: parameters are abundant, so understanding cannot mean tracking each weight individually. Which collective quantities describe the network's behaviour, and how can we derive them from its microscopic model?

Mean-field theory gives one such effective description: instead of tracking every neuron and every random weight, one tracks a small number of self-consistent fields. In the signal-propagation version used here, those fields are variances, covariances, and correlations. This is the deep information propagation line of mean-field theory \cite{poole2016exponentialexpressivitydeepneural,schoenholz2017deepinformationpropagation}, with broader background in statistical-mechanics treatments of deep learning \cite{Bahri2020StatMechDeepLearning,bahri2024leshoucheslecturesdeep}. The same infinite-width covariance recursion can also be read as the NNGP kernel recursion \cite{Lee2018DNNasGP}; we use the MFT language because our questions concern fixed points, susceptibilities, correlation lengths, and critical exponents rather than the induced function-space prior or training dynamics.

Using the signal-propagation setup of Sec.~\ref{sec:mft_background}, consider a depth-$L$ MLP with activation $\phi:\mathbb R\to\mathbb R$, whose weights and biases are initialized as
\begin{equation}
W^{l}_{ij}\sim \mathcal{N}\left(0,\frac{\sigma_{w}^{2}}{N_{l-1}}\right), \qquad b^{l}_{i}\sim \mathcal{N}\left(0,\sigma_{b}^{2}\right),
\end{equation}
and propagated by
\begin{equation}
z^{l}_{i;a}= \sum_{j=1}^{N_{l-1}} W^{l}_{ij} y^{l-1}_{j;a}+b^{l}_{i}, \qquad y^{l}_{i;a}=\phi\left(z^{l}_{i;a}\right),
\end{equation}
where $W^l\in\mathbb R^{N_l\times N_{l-1}}$, $y^0=x$, and $a$ labels the input. We write
\begin{equation}
\int Dz (\cdots)=\frac{1}{\sqrt{2\pi}}\int_{-\infty}^{\infty}dz e^{-z^{2}/2}(\cdots),
\end{equation}
and use the same convention for multiple Gaussian variables.

The Gaussian closure comes from the fact that each preactivation is a sum of many independently initialized weighted inputs. In the infinite-width limit this central-limit effect becomes exact layer-by-layer, while finite-width corrections are organized perturbatively in the depth-to-width aspect ratio, schematically $L/N$ \cite{roberts2022principlesdeeplearningtheory}. For $l\ge2$, the single-input variance and two-input covariance recursions are the closed equations already quoted in Sec.~\ref{sec:mft_background}:
\begin{align}
q^{l}_{aa}&=\sigma_{w}^{2}\int Dz \phi^{2}\left(\sqrt{q^{l-1}_{aa}} z\right)+\sigma_{b}^{2},\nonumber\\
q^{l}_{ab}&=\sigma_{w}^{2}\int Dz_{1} Dz_{2} \phi(u_{1}) \phi(u_{2})+\sigma_{b}^{2},\nonumber\\
u_{1}&=\sqrt{q^{l-1}_{aa}} z_{1},\qquad c^{l-1}_{ab}=\frac{q^{l-1}_{ab}}{\sqrt{q^{l-1}_{aa} q^{l-1}_{bb}}},\nonumber\\
u_{2}&=\sqrt{q^{l-1}_{bb}}\left(c^{l-1}_{ab}z_{1}+\sqrt{1-\left(c^{l-1}_{ab}\right)^{2}} z_{2}\right).
\end{align}
Once the variance relaxes to $q^*$, these equations induce a one-dimensional correlation map $c^l_{ab}=F(c^{l-1}_{ab})$. Without dropout, perfect alignment is a fixed point, $F(1)=1$. Its linear stability is controlled by the angular susceptibility
\begin{equation}
\chi_{1}=\left.\frac{\partial c^{l}_{ab}}{\partial c^{l-1}_{ab}}\right|_{c_{ab}=1}=\sigma_{w}^{2}\int Dz \big[\phi'(\sqrt{q^{*}} z)\big]^{2},
\end{equation}
called $\chi_\perp$ in the main text and in \cite{roberts2022principlesdeeplearningtheory}. The ordered, chaotic, and critical regimes correspond to $\chi_1<1$, $\chi_1>1$, and $\chi_1=1$ respectively; the last case is the edge-of-chaos, where the cross-input correlation length can diverge.

The usual He initialization follows from this criterion in the special case of \relu{}. Since $\phi'(z)=\theta(z)$ almost everywhere,
\begin{equation}
\int Dz \big(\phi'(\sqrt{q^{*}}z)\big)^2=\frac12,
\qquad \chi_1=1 \Longrightarrow \sigma_w^2=2,
\end{equation}
which is the variance-$2/N$ initialization. For \relu{}, the variance recursion is
\begin{equation}
q^{l}_{aa}=\frac{\sigma_w^2}{2} q^{l-1}_{aa}+\sigma_b^2,
\end{equation}
so with nonzero bias the finite-$q^*$ condition requires $\sigma_w^2<2$, while exact angular criticality sets $\sigma_w^2=2$. In practice one either sets biases to zero or works slightly subcritical to keep the variance scale finite.

Finally, the correlation length used throughout the paper is obtained by linearizing the correlation map around its fixed point. With $r_c=F'(c^*)$ and $0<|r_c|<1$, small deviations obey
\begin{equation}
|c^*-c^l_{ab}|\propto e^{-l/\xi_c},\qquad \xi_c^{-1}=-\log|r_c|.
\end{equation}
The analogous variance length $\xi_q$ tracks relaxation of $q^l_{aa}$ to $q^*$, but the edge-of-chaos phenomenon is primarily controlled by $\xi_c$: it measures how many layers preserve distinctions between different inputs. This is the scale deformed by dropout in the main analysis.
\FloatBarrier

\subsection{Derivation of Correlation Map and Derivatives with Dropout}\label{app:price_dropout}

Fix the positive variance \(\bar q^*\) in the presence of dropout, given by \eqref{eq:qstar-dropout}, and let \((u_1,u_2)\) be centered Gaussian with marginal variance \(\bar q^*\) and correlation \(c\). Independent inverted-dropout masks across inputs give the normalized map
\begin{equation}
\bar F_\rho(c)=\frac{\sigma_w^2\mathbb E_c[\phi(u_1)\phi(u_2)]+\sigma_b^2}{\bar q^*},
\qquad h=1-\bar F_\rho(1),
\label{eq:Frho_def_app}
\end{equation}
with \(h\) given by \eqref{eq:Delta_exact}.

Assume \(\phi\in C^2\) and \(\phi,\phi',\phi''\in L^2(\mathcal N(0,\bar q^*))\). Price's theorem \cite{Price1958GaussianInputs,Voigtlaender2021Price}, applied at fixed \(\bar q^*\), states that for \(|c|<1\),
\begin{equation}
\frac{\mathrm d^k}{\mathrm dc^k}\mathbb E_c[\phi(u_1)\phi(u_2)]
=(\bar q^*)^k\mathbb E_c[\phi^{(k)}(u_1)\phi^{(k)}(u_2)],
\qquad k=1,2.
\end{equation}
Strong continuity of Gaussian correlation operators in \(L^2\) then justifies \(c\to1^-\), giving the one-sided derivatives
\begin{align}
\bar F_\rho'(1^-)&=\sigma_w^2\int Dz\,[\phi'(\sqrt{\bar q^*}z)]^2=\chi_\rho,\\
\bar F_\rho''(1^-)&=\sigma_w^2\bar q^*\int Dz\,[\phi''(\sqrt{\bar q^*}z)]^2=g_\rho,
\end{align}
as used in Eqs.~\eqref{eq:chi_dropout}--\eqref{eq:g_dropout}. Taking \(c\to1^-\) requires the stated Gaussian square-integrability conditions; the theorem for nondegenerate \(|c|<1\) alone does not establish finite derivatives at perfect alignment.

\FloatBarrier

\subsection{Extensions to Infinite-Channel CNNs and ResNets}
\label{app:cnn_extension}

Although our derivation is for MLPs, the mechanism that determines the universality class is tied to the local Gaussian kernel rather than to the MLP architecture itself. Other architectures introduce additional structure that can change the exact exponents, but the qualitative smooth-versus-kinked distinction should remain when the same local channel controls the recursion at infinite width.

For example, the same local channel which appears in MLPs re-emerges in infinite-width CNNs when the number of channels is taken to infinity: preactivations at different spatial positions become jointly Gaussian across channels, and the covariance recursion composes the same bivariate Gaussian activation expectation with a linear operator that averages over convolutional filter offsets \cite{cnnslecun,xiao2018dynamicalisometrymeanfield}. In schematic form, a spatial covariance \(K^\ell_{\alpha\beta}(x,x')\) evolves as \(K^{\ell+1}=\sigma_b^2+\sigma_w^2\mathcal A[V_\phi(K^\ell)]\), where \(V_\phi\) is the scalar Gaussian channel analyzed above and \(\mathcal A\) is the convolutional averaging operator. Independent dropout shifts the single-input variance entering \(V_\phi\) as in the fully connected case, while the cross-input covariance is still controlled by the same bivariate expectation. Therefore, for an isolated leading spatial mode, convolution changes the relevant eigenvalue and mode shape but not the local smooth-versus-kinked non-analyticity: smooth activations retain the analytic Landau equation, while \relu{}-like kinks retain the \(m^{3/2}\) branch point. A complete CNN theory would promote \(m\) to a spatial covariance-mode amplitude and track boundary conditions, pooling, and degeneracies of \(\mathcal A\).

In ResNets, skip connections can change exponential convergence to subexponential or polynomial behavior, and norms may drift rather than relax to a fixed point \cite{yang2017meanfieldresnets}. Still, the residual branch evaluates the same nonlinear Gaussian channel \(V_\phi\), so the local analytic distinction is unchanged. The tanh/\relu{} asymptotic split observed in \cite{yang2017meanfieldresnets} is consistent with our smooth/kinked classes; a dropout-deformed ResNet theory would additionally track residual-branch strength and norm drift.

\FloatBarrier

\subsection{Regularization Reach}
\label{app:regularization_reach}

For the comparison leading to \eqref{eq:regularization_reach}, write \(\tilde y_k^{(\ell)}\) for the trajectory masked only at layer \(\ell\) and \(y_k\) for the clean trajectory. The normalized second-moment overlap and its decorrelation are
\begin{equation}
c_k^{(\ell)}
\equiv
\frac{\mathbb E_p[\langle \tilde y_k^{(\ell)},y_k\rangle]}
{\sqrt{\mathbb E_p[\|\tilde y_k^{(\ell)}\|^2]\|y_k\|^2}},
\qquad \eta_k^{(\ell)}=1-c_k^{(\ell)}.
\end{equation}
Here \(\mathbb E_p\) averages over the mask at fixed clean activations, with \(\|y_k\|>0\); the normalization uses mask-averaged second moments. The main text derives the initial overlap change, its weak-field relation from \eqref{eq:Delta_weakdrop}, and its downstream propagation.

\begin{figure}[t]
\centering
\includegraphics[width=\linewidth]{\PaperRoot/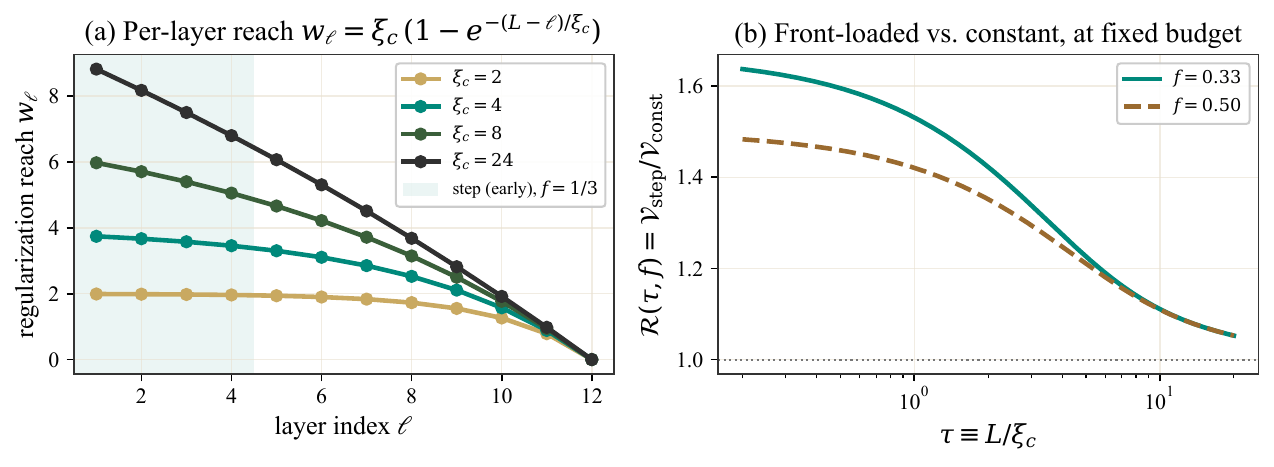}
\caption{Weights for regularization reach at fixed network depth \(L=12\). (a)~Per-layer weight \(w_\ell=\xi_c(1-e^{-(L-\ell)/\xi_c})\) is monotone decreasing in \(\ell\) for any \(\xi_c\); the linear program over the dropout budget therefore fills the earliest layers first (shaded: early step with \(f=1/3\)). For \(L\ll\xi_c\) the reach is essentially linear in \(L-\ell\); for \(L\gg\xi_c\) it saturates at \(\xi_c\). (b)~Ratio \(\mathcal R(\tau,f)=\mathcal V_{\rm step}/\mathcal V_{\rm const}\) of front-loaded to constant schedules at fixed budget, with \(\tau\equiv L/\xi_c\). Here $\mathcal V$ denotes summed exposure at a common background $\xi_c$. In this continuum comparison the ratio approaches $2-f$ as $\tau\to0$ and one as $\tau\to\infty$.}
\label{fig:regularization_reach}
\end{figure}

For the multiplier \(0<\lambda<1\) of the common ordered background used there, the exact discrete weight in \eqref{eq:regularization_reach} is
\begin{equation}
W_\ell=\sum_{r=1}^{L-\ell}\lambda^r
=\frac{\lambda(1-\lambda^{L-\ell})}{1-\lambda}
=\frac{\lambda}{\xi_c(1-\lambda)}w_\ell,
\qquad \xi_c=-\frac{1}{\log\lambda}.
\end{equation}
Thus the continuum weight \(w_\ell\) differs by a factor that does not depend on position and gives the same ordering (Fig.~\ref{fig:regularization_reach}). The perturbation at readout instead scales as \(h_\ell\lambda^{L-\ell}\), which favors later masks; front-loading follows from summed exposure across layers.

\begin{samepage}
This approximation requires weak masks, neglected variance transients, and common \(C\) and \(\xi_c\) across mask positions. At an exactly marginal clean channel, \(\lambda=1\), higher-order terms produce algebraic relaxation; the limit \(W_\ell\to L-\ell\) describes the linearized finite-depth response, not that nonlinear relaxation. Adding exposures from several masks is a further assumption, which mask interactions and training may alter. Its relation to generalization remains an empirical question.
\end{samepage}

In the propagation approximation, rearranging the same saturated layers leaves the cost unchanged. At finite depth, its ordering is fixed by
\begin{equation}
\max_{\{h_\ell\}}\sum_{\ell=1}^L h_\ell W_\ell
\qquad
\text{s.t.}
\qquad
\sum_{\ell=1}^L h_\ell=L\bar h,
\qquad
0\le h_\ell\le h_{\max}.
\end{equation}
Since \(W_1>W_2>\cdots>W_L\), transferring budget from a later active layer to an earlier unsaturated one increases exposure. The solution therefore fills the earliest layers first:
\begin{equation}
h_\ell^\star=
\begin{cases}
h_{\max}, & 1\le \ell\le k,\\
r, & \ell=k+1\le L,\\
0, & \ell>k+1,
\end{cases}
\qquad
k=\left\lfloor\frac{L\bar h}{h_{\max}}\right\rfloor,
\quad r=L\bar h-kh_{\max}.
\end{equation}
For \(h_{\max}>0\) and \(0\le\bar h\le h_{\max}\), the residual \(r\in[0,h_{\max})\) enforces the budget, with at most one partially filled layer. Sliding a fixed saturated block \(B\) toward later layers reduces \(\sum_{\ell\in B}W_\ell\), providing a direct empirical test of the reach heuristic.

\FloatBarrier

\section{Critical Exponents and Non-Analyticity}
\label{ap:exponent_derivations}

The Hermite decomposition in App.~\ref{sec:hermite} determines which expansion is valid near perfect alignment. In the Gaussian channel,
\begin{equation}
F(c)=\frac{\sigma_w^2\sum_{n\ge0}a_n^2 c^n+\sigma_b^2}{q^*}.
\end{equation}
The derivative moments assumed for smooth activations make this series differentiable to the order needed at $c=1$, giving the finite Taylor expansion used in the Landau equation below. This does not require analyticity or convergence of an infinite Taylor series. A kink produces a power-law Hermite tail, so high-degree modes contribute collectively as \(c\to1\); the Taylor expansion is replaced by the branch-point term \(m^{3/2}\), which is the source of the kinked exponents.

\begin{table}[h]
\caption{Definitions of critical exponents near the correlation edge in the infinite-width theory ($t,h\to0$). Here $C_{\rm sing}=-\partial_t^2 f_{\rm sing,on}(t,0)$, where $f_{\rm sing,on}$ is the singular Landau potential evaluated at its stable stationary point.}
\label{tab:math_exponents}
\centering
\renewcommand{\arraystretch}{1.8}
\begin{tabular}{lc}
\toprule
Exponent & Mathematical Definition \\
\midrule
$\beta$ & $m \sim t^\beta \quad (h=0,\ t\to0^+)$ \\
$\delta$ & $m \sim |h|^{1/\delta} \quad (t=0)$ \\
$\gamma$ & $\chi_M \equiv \left. \frac{\partial m}{\partial h} \right|_{h \to 0} \sim |t|^{-\gamma}$ \\
$\alpha$ & $C_{\rm sing} \sim t^{-\alpha} \quad (t\to0^+)$ \\
$\nu_t$ & $\xi \sim |t|^{-\nu_t} \quad (h=0)$ \\
$\nu_\rho$ & $\xi \sim |h|^{-\nu_\rho} \quad (t=0)$ \\
$\theta_{\rm rel}$ & $m \sim \ell^{-\theta_{\rm rel}} \quad (t=0, h=0)$ \\
\bottomrule
\end{tabular}
\end{table}

\subsection{Smooth Networks}

The nonnegative solution of the leading smooth equation is
\begin{equation}
m(t,h)=\frac{t+\sqrt{t^2+2g_\rho h}}{g_\rho},
\label{eq:m_smooth_solution}
\end{equation}
and at the edge,
\begin{equation}
m(0,h)=\sqrt{\frac{2h}{g_\rho}}.
\label{eq:m_at_edge}
\end{equation}

Along the critical isotherm \(t=0\), \eqref{eq:EoS_correct} gives
\begin{equation}
h=\frac{g_\rho}{2}m^2 \qquad\Rightarrow\qquad m\sim h^{1/2},
\end{equation}
so \(\delta=2\). The analogue of the magnetic susceptibility is \(\chi_M(t)\equiv \left.\partial m/\partial h\right|_{h\to 0}\).
On the subcritical side \(t<0\), the analytic branch satisfies \(m\simeq -h/t\), hence \(\chi_M(t)\sim |t|^{-1}\) and \(\gamma=1\).
Along \(h=0\), the physical branch is \(m=0\) for \(t\le 0\) and \(m=2t/g_\rho\) for \(t\ge 0\), so \(m\sim t\) as \(t\to 0^+\) and \(\beta=1\).

\vspace{0.25em}
\noindent
The depth correlation length follows from linearizing the correlation dynamics about the displaced fixed point.
To leading order,
\begin{equation}
\lambda \equiv \bar F_\rho'(c^*) = \bar F_\rho'(1-m)\simeq \chi_\rho-g_\rho m = 1+t-g_\rho m,
\end{equation}
and substituting \eqref{eq:m_smooth_solution} gives
\begin{equation}
\lambda(t,h)=1-\sqrt{t^2+2g_\rho h}.
\label{eq:lambda_smooth_correct}
\end{equation}
For \(\lambda\simeq 1\) with \(\lambda>0\), the depth correlation length satisfies
\begin{equation}
\xi_{c,\rho}^{-1}\equiv -\log\lambda \simeq 1-\lambda = \sqrt{t^2+2g_\rho h}.
\end{equation}
Thus \(\xi_{c,\rho}(t,0)\sim |t|^{-1}\) and \(\nu_t=1\), while \(\xi_{c,\rho}(0,h)\sim h^{-1/2}\) and \(\nu_\rho=1/2\).
Equivalently, \(y_t=1/\nu_t=1\) and \(y_\rho=1/\nu_\rho=2\), so dropout is the more relevant perturbation of the correlation edge.

At the marginal point \(t=h=0\), the same smooth normal form gives the depth relaxation law:
\begin{equation}
m_{\ell+1}-m_\ell\simeq-\frac{g_\rho}{2}m_\ell^2.
\end{equation}
In a continuum-depth approximation this gives \(m(\ell)\sim \ell^{-1}\), hence \(\theta_{\rm rel}=1\) for the smooth class.

\vspace{0.25em}
\noindent
A pseudo-free-energy functional consistent with \eqref{eq:EoS_correct} is
\begin{equation}
f(t,m;h) = -\frac{t}{2} m^2 + \frac{g_\rho}{6} m^3 - h m,
\label{eq:free_energy_correct}
\end{equation}
since \(\partial f/\partial m=0\) is equivalent to \eqref{eq:EoS_correct}. As with a low-order one-particle-irreducible (1PI) effective action, its stationarity condition gives the equation of state, while its singular part evaluated at the stationary point gives analogues of the thermodynamic exponents; analytic \(m\)-independent backgrounds can be added without changing the physics of the recursion.

At \(h=0\), substituting the physical branch
\(m=0\) for \(t\le 0\) and \(m=2t/g_\rho\) for \(t\ge 0\) yields
\begin{equation}
f_{\rm on}(t,0)=0 \quad (t\le 0), \qquad f_{\rm on}(t,0)=-\frac{2}{3g_\rho^{2}} t^{3}\quad (t\ge 0).
\end{equation}
Therefore the analogue of the specific heat, \(C\equiv -\partial_t^{2} f_{\rm on}\), stays finite and vanishes linearly:
\begin{equation}
C(t)=\frac{4}{g_\rho^{2}} t \Theta(t)\sim |t|,
\end{equation}
where $\Theta(t)$ is the Heaviside step function. In standard exponent language \(C_{\rm sing}(t)\sim |t|^{-\alpha}\), this corresponds to
\begin{equation}
\alpha=-1,
\end{equation}
up to an additive analytic background \(f_0(t,h)\) that does not affect the equation of state.

Collecting exponents for the smooth universality class:
\begin{equation}
\beta=1, \qquad \gamma=1, \qquad \delta=2, \qquad \nu_t=1, \qquad \nu_\rho=\frac{1}{2}, \qquad \theta_{\rm rel}=1, \qquad \alpha=-1.
\end{equation}

\subsection{Kinked Networks}
\label{sec:kinks}

For \relu{}, take \(\sigma_b=0\) and the same inverted-dropout convention, with masks drawn independently across inputs. The clean correlation map has the arc-cosine form \cite{ChoSaul2009ArcCosine}
\begin{align}
F_{\ReLU}(c) &= \frac{1}{\pi} \Bigl[ \sqrt{1-c^{2}} + \bigl(\pi-\cos^{-1}c\bigr)c \Bigr].
\label{eq:relu_map_rho1}
\end{align}
Near perfect alignment, setting \(c=1-m\) with \(0<m\ll 1\), one obtains the non-analytic expansion
\begin{align}
F_{\ReLU}(1-m) &= 1-m+\kappa m^{3/2}+{\cal O}(m^{2}),
\label{eq:relu_expand}\\
\kappa &\equiv \frac{2\sqrt{2}}{3\pi}. \nonumber
\end{align}
With dropout, the field is exactly
\begin{equation}
h \;=\; 1-\bar F_\rho(1) \;=\; 1-\rho,
\label{eq:relu_h_correct}
\end{equation}
since independent masks supply the mismatch at \(c=1\). For the literal bias-free \relu{} map, \(\bar F_\rho(c)=\rho F_{\ReLU}(c)\), so \(\chi_\rho=\rho\) and \(t=\rho-1=-h\). Thus the \(t=0,h>0\) direction below is the field direction of the kinked normal form, while standard \relu{} dropout follows a constrained path through the same scaling function.

The fixed-point condition \(c^*=\bar F_\rho(c^*)\), with \(m^*=1-c^*\ll1\), combines the field, slope and branch term into
\begin{equation}
h \;=\; \kappa m^{3/2} - t m + \mathcal O(m^2), \qquad t\equiv \chi_\rho-1.
\label{eq:eos_kink_correct}
\end{equation}
Along the critical isotherm \(t=0\) of the normal form, this gives
\begin{equation}
m^* \;\sim\; \left(\frac{h}{\kappa}\right)^{2/3}, \qquad (t=0),
\label{eq:kink_mstar}
\end{equation}
so the smooth exponent \(\delta=2\) is replaced by
\begin{equation}
m \propto h^{1/\delta_{\mathrm{kink}}}, \qquad \delta_{\mathrm{kink}}=\frac{3}{2}.
\label{eq:kink_delta}
\end{equation}
The change from $\delta=2$ to $\delta=3/2$ separates smooth and kinked activations into different mean-field universality classes. The depth scale follows from linearizing the correlation dynamics around \(c^*\). Differentiating \eqref{eq:relu_expand} gives
\begin{equation}
\lambda \equiv \bar F_\rho'(c^*) \simeq 1+t-\frac{3\kappa}{2}\sqrt{m^*}.
\label{eq:kink_slope}
\end{equation}
At \(t=0\) we therefore have
\begin{equation}
1-\lambda \propto \sqrt{m^*}\propto h^{1/3}.
\end{equation}
Using \(\xi_{c,\rho}^{-1}\simeq 1-\lambda\) for \(\lambda\simeq 1\) yields
\begin{equation}
\xi_{c,\rho}(0,h) \sim h^{-1/3}, \qquad \nu_{\rho,\mathrm{kink}}=\frac{1}{3},
\label{eq:kink_nu}
\end{equation}
to be compared with the smooth result \(\xi_{c,\rho}(0,h)\sim h^{-1/2}\).
For ordinary bias-free \relu{} dropout, substituting the constrained path \(t=-h\) into \eqref{eq:eos_kink_correct} gives
\begin{equation}
h=\kappa m^{3/2}+hm+\mathcal O(m^2).
\end{equation}
Since the leading solution has \(m\sim h^{2/3}\), the \(hm\) term is subleading. Moreover,
\begin{equation}
1-\lambda\simeq h+\frac{3\kappa}{2}\sqrt m\sim h^{1/3},
\end{equation}
so the same \(\nu_{\rho,\mathrm{kink}}=1/3\) survives along this constrained path.

The remaining kinked exponents follow from the same equation of state. At \(h=0\) and \(t>0\),
\begin{equation}
\kappa m^{3/2}=tm \qquad\Rightarrow\qquad m=\left(\frac{t}{\kappa}\right)^2,
\end{equation}
so \(\beta_{\rm kink}=2\). On the subcritical side \(t<0\), the small-field balance is \(h\simeq |t|m\), giving \(\chi_M\sim |t|^{-1}\) and therefore \(\gamma_{\rm kink}=1\). Substituting the zero-field branch into \eqref{eq:kink_slope} gives \(1-\lambda\propto t\), so \(\xi_{c,\rho}(t,0)\sim t^{-1}\) and \(\nu_{t,\rm kink}=1\).

Adding nonzero \relu{} bias variance does not restore a finite-\(q^*\) critical isotherm. With inverted dropout,
\begin{equation}
\bar q_{\ell+1}=\frac{\sigma_w^2}{2\rho}\bar q_\ell+\sigma_b^2,
\end{equation}
so finite variance requires \(a\equiv\sigma_w^2/(2\rho)<1\), while angular criticality requires \(\chi_\rho=\sigma_w^2/2=1\), incompatible with \(\rho<1\). Equivalently, \(h=a(1-\rho)\) and \(t=\rho a-1=-h-(1-a)\). Fixed subcritical detuning cuts off the divergence as \(h\to0\), whereas \(1-a=\mathcal O(h^{1/3})\) retains the \(h^{-1/3}\) scaling with a prefactor set by the rescaled detuning; \(1-a=o(h^{1/3})\) recovers the critical-isotherm prefactor.

The pseudo-free-energy corresponding to \eqref{eq:eos_kink_correct} is
\begin{equation}
f_{\mathrm{kink}}(t,m;h) \;=\; -\frac{t}{2} m^2 + \frac{2\kappa}{5} m^{5/2} - h m,
\label{eq:free_energy_kink}
\end{equation}
since $\partial f_{\mathrm{kink}}/\partial m=0$ gives $h=\kappa m^{3/2}-tm$ at leading order. Integrating the $m^{3/2}$ term in the equation of state gives the $m^{5/2}$ term in this potential.\footnote{
As before, $f_{\mathrm{kink}}$ is defined only up to addition of an $m$-independent analytic background $f_0(t,h)$, which does not affect the equation of state.
}
At zero field the physical branch is $m=0$ for $t\le 0$ and $m=(t/\kappa)^2$ for $t\ge 0$, so the on-shell free energy behaves as
\begin{equation}
f_{\mathrm{kink,on}}(t,0)=0 \quad (t\le 0), \qquad f_{\mathrm{kink,on}}(t,0)=-\frac{1}{10\kappa^{4}} t^{5}\quad (t\ge 0).
\end{equation}
Thus the analogue of the specific heat, $C\equiv-\partial_t^{2} f_{\mathrm{kink,on}}$, stays finite and in fact vanishes as
\begin{equation}
C(t)\;=\;\frac{2}{\kappa^{4}} t^{3} \Theta(t)\;\sim\;|t|^{3}.
\end{equation}
In the standard exponent convention $C_{\mathrm{sing}}(t)\sim |t|^{-\alpha}$, this corresponds to
\begin{equation}
\alpha_{\mathrm{kink}}=-3.
\end{equation}

At the marginal point \(\rho=1\), \(t=0\), the same non-analytic term controls relaxation. Writing \(m_\ell\equiv1-c_\ell\), \eqref{eq:relu_expand} gives
\begin{equation}
m_{\ell+1}-m_\ell \simeq -\kappa m_\ell^{3/2}.
\label{eq:kink_relax}
\end{equation}
In a continuum-depth approximation \(m_{\ell+1}-m_\ell\to \partial_\ell m\), we obtain
\begin{equation}
\partial_\ell m \simeq -\kappa m^{3/2} \qquad\Rightarrow\qquad m(\ell)\sim \ell^{-2},
\label{eq:kink_relax_sol}
\end{equation}
which reproduces the characteristic \(1-c_\ell\sim \ell^{-2}\) relaxation of \relu{} activations at the edge-of-chaos \cite{hayou2019impact} and gives \(\theta_{\rm rel}=2\).

\subsection{Origin of the $m^{3/2}$ Term for Kinked Activations}
\label{app:tube_argument_kink}

To establish the fractional power beyond \relu{}, assume \(\phi\) is continuous and piecewise \(C^2\), with a finite slope jump at the kink, a locally bounded regular second derivative, and Gaussian square-integrability of \(\phi\), \(\phi'\) and that regular derivative. Absorb the fixed marginal variance into \(\phi\) and let \((u_1,u_2)\) be standard Gaussian marginals with correlation \(c=1-m\), \(0<m\ll1\). Introduce the sum and difference coordinates
\begin{equation}
s=\frac{u_{1}+u_{2}}{\sqrt{2}}, \qquad d=\frac{u_{1}-u_{2}}{\sqrt{2}}.
\end{equation}
A direct covariance computation gives
\begin{equation}
\mathrm{Var}(s)=1+c=2-m, \qquad \mathrm{Var}(d)=1-c=m, \qquad \mathbb{E}[sd]=0.
\end{equation}
The transverse fluctuation therefore has root-mean-square size \(\sqrt m\), concentrating the joint Gaussian measure in a thin tube around \(u_1=u_2\). The probability of straddling the kink is \(\mathcal O(\sqrt m)\), and within that region the activation mismatch produced by the slope jump has root-mean-square size \(\mathcal O(\sqrt m)\). The kernel correction is consequently \(\mathcal O(\sqrt m)\times\mathcal O(m)=\mathcal O(m^{3/2})\).

For an explicit calculation with a single kink, put it at \(u=0\), let its one-sided slopes be \(a_\pm\), and write locally
\begin{equation}
\phi(u)=\phi(0)+a_- u+\big(a_+-a_-\big) \ReLU(u)+r(u), \qquad r(u)=\mathcal O(u^2)\ \ \text{as }u\to 0,
\end{equation}
where \(r\) has no slope jump. For \(K(c)=\mathbb E[\phi(u_1)\phi(u_2)]\), Price's second derivative pairs the delta functions from the jump \(\Delta a=a_+-a_-\), giving
\begin{equation}
K''(c)=\frac{(\Delta a)^2}{2\pi\sqrt{1-c^2}}+\mathcal O(1).
\end{equation}
The mixed and regular terms are bounded under the stated assumptions. Integrating twice toward \(c=1\) yields
\begin{equation}
K(1-m)=K(1)-K'(1)m+\frac{\sqrt2(\Delta a)^2}{3\pi}m^{3/2}+\mathcal O(m^2).
\end{equation}
At the marginal zero-field point, variance normalization therefore gives
\begin{equation}
F(1-m)=1-m+\kappa_\phi m^{3/2}+\mathcal O(m^2).
\end{equation}
For \relu{}, \(F=2K\) recovers \(\kappa=2\sqrt2/(3\pi)\) in \eqref{eq:relu_expand}. Moving the kink changes the Gaussian density at its location and hence the amplitude, while the exponent remains \(3/2\). Mere continuity of \(r'\) would not suffice for the stated remainder.

\subsection{Crossover Functions and Limiting Branches}
\label{app:crossover_details}

The smooth solution \eqref{eq:m_smooth_solution} yields the rescaled curve \eqref{eq:universal_scaling_function_smooth_correct}. Its crossover scale is
\begin{equation}
|t|\sim\sqrt{g_\rho h}.
\label{eq:crossover_scale_smooth}
\end{equation}
For $|t|\ll\sqrt{g_\rho h}$, dropout controls the response,
\begin{equation}
m\sim\sqrt{\frac{2h}{g_\rho}}.
\label{eq:field_dominated_smooth}
\end{equation}
For $|t|\gg\sqrt{g_\rho h}$, the two limiting branches are
\begin{align}
t<0:\qquad m&=\frac{h}{|t|}+\mathcal O\!\left(\frac{g_\rho h^2}{|t|^3}\right),\\
t>0:\qquad m&=\frac{2t}{g_\rho}+\frac{h}{t}+\mathcal O\!\left(\frac{g_\rho h^2}{t^3}\right).
\end{align}
These are asymptotic expansions of the leading equation of state, with its coefficients held fixed; the full smooth map also contains $\mathcal O(m^3)$ terms.

For kinked activations, define the rescaled detuning $u$ and response function $\mathcal F$ by
\begin{align}
u&\equiv\frac{t}{\kappa^{2/3}h^{1/3}},
\label{eq:def_u}\\
m(t,h)&=\left(\frac{h}{\kappa}\right)^{2/3}\mathcal F(u).
\label{eq:m_kink_scaling_form}
\end{align}
The leading equation \eqref{eq:eos_kink_correct} fixes $\mathcal F$ through the positive real root $y(u)$ of
\begin{equation}
y^3-uy^2-1=0,\qquad \mathcal F(u)=y(u)^2.
\label{eq:cubic_y_correct}
\end{equation}
The crossover occurs at
\begin{equation}
|t|\sim\kappa^{2/3}h^{1/3}.
\label{eq:crossover_scale_kink}
\end{equation}
The ordinary bias-free \relu{} path $t=-h$ satisfies $u\to0$ as $h\to0$, retaining the field-dominated $\xi\sim h^{-1/3}$ law derived above. Corrections to this collapse include the $\mathcal O(m^2)$ terms in the kinked map, changes in its coefficients away from the reference edge, and the nonlinear microscopic relation between $h$ and $\rho$.

\section{Hermite Spectral Perspective}
\label{sec:var-opt-act}
\label{sec:hermite}

\begin{figure}[ht]
\centering
\includegraphics[width=0.58\linewidth]{\PaperRoot/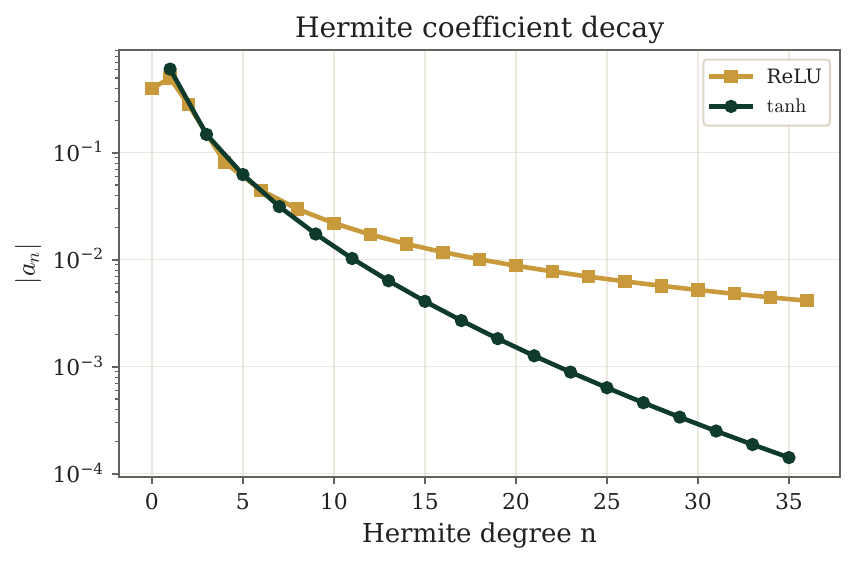}
\caption{Hermite coefficient magnitudes for the unscaled activations $\ReLU(z)$ and $\tanh(z)$. The \relu{} coefficients use their exact expression; tanh uses 220-point Gauss--Hermite quadrature. The coefficient tail exposes the difference in regularity at $c=1$.}
\label{fig:hermite-decomposition}
\end{figure}

The distinction between smooth and kinked activations can also be read from their Hermite coefficients. For the channel without dropout, at a positive variance fixed point $q^*$, set $f(z)=\phi(\sqrt{q^*}z)$ and assume $f\in L^2(Dz)$. The normalized probabilists' Hermite polynomials $h_n(z)=\mathrm{He}_n(z)/\sqrt{n!}$ form an orthonormal basis under the standard Gaussian measure, so
\begin{equation}
f(z)=\sum_{n\ge0}a_n h_n(z),\qquad a_n=\int Dz\,f(z)h_n(z),\qquad \sum_{n\ge0}a_n^2=\int Dz\,f(z)^2.
\end{equation}
For jointly standard Gaussian variables of correlation $c$, this basis satisfies $\mathbb E[h_n(z_1)h_m(z_2)]=c^n\delta_{nm}$; using Lemma~11 of \citet{Daniely2016DualView}, the correlation map becomes
\begin{equation}
F(c)=\frac{\sigma_w^2\sum_{n\ge0}a_n^2c^n+\sigma_b^2}{q^*},\qquad q^*=\sigma_w^2\sum_{n\ge0}a_n^2+\sigma_b^2.
\end{equation}
The coefficient tail determines whether the derivatives needed near perfect alignment remain finite: for each integer $k\ge1$ whose weighted sum converges,
\begin{equation}
F^{(k)}(1)=\frac{\sigma_w^2}{q^*}\sum_{n\ge k}\frac{n!}{(n-k)!}a_n^2.
\end{equation}
Thus the Gaussian derivative moments assumed for the smooth class give the finite-order Taylor expansion used in the Landau equation; convergence of an infinite Taylor series is unnecessary. The smooth example $\tanh$ has rapidly decaying coefficients, whereas \relu{} has a power-law tail for which $F''(1)$ diverges (Fig.~\ref{fig:hermite-decomposition}). A kink with a finite jump in slope, under the regularity assumptions in App.~\ref{app:tube_argument_kink}, produces the $m^{3/2}$ correction in place of the smooth class's $m^2$ term.

\FloatBarrier

\section{Experimental Details and Additional Figures}
\label{app:experimental_info}
This appendix gives the training settings and additional figures. Curves and tables comparing all profiles use the first set of seeds; Table~\ref{tab:loss_improvements} adds seven fresh pairs for the selected \relu{} comparison and five for the ViT with dropout in both blocks, bringing each to ten pairs while keeping the settings fixed. The notebooks and saved results, including the additional runs, are available in the \href{https://github.com/luklacasito/icml}{\texttt{icml}} repository.

\subsection{Mean-Field Fits and Scaling Collapse}
Table~\ref{tab:mf_vs_fit_exponents} reports fits to the numerical mean-field curves. The repository includes the raw arrays, fitting windows, and a CPU script that regenerates Fig.~\ref{fig:dropout_scaling}. Tanh expectations use 60-point Gauss--Hermite quadrature. The \relu{} calculation uses the formal channel $F(c)=\chi F_{\ReLU}(c)+1-\chi/\rho$, where $F_{\ReLU}$ is the normalized arc-cosine map. Thus $t=\chi-1$ and $h=\chi(1/\rho-1)$. This family probes the local normal form; its independently tuned parameters need not correspond to nonnegative bias variance in a finite-variance \relu{} network.
\begin{table}[ht]
\caption{Critical exponents from log--log fits in Fig.~\ref{fig:dropout_scaling}. The $\pm$ values are least-squares fit standard errors, not seed uncertainty or bounds on numerical error. Deviations from the asymptotic predictions reflect the finite fitting windows and numerical approximation; small fit errors do not remove these systematic effects.}
\label{tab:mf_vs_fit_exponents}
\centering
\begin{tabular}{lcc}
\toprule
Exponent & Numerical fit & Asymptotic prediction \\
\midrule
$\nu$ & $1.0190 \pm 0.0014$ & $1$ \\
$\beta_{\mathrm{tanh}}$ & $0.9879 \pm 0.0007$ & $1$ \\
$\beta_{\ReLU}$ & $1.9642 \pm 0.0019$ & $2$ \\
$\theta_{\mathrm{rel,tanh}}$ & $1.00483 \pm 0.00002$ & $1$ \\
$\theta_{\mathrm{rel},\ReLU}$ & $1.98697 \pm 0.00005$ & $2$ \\
$1/\delta_{\mathrm{tanh}}$ & $0.5171 \pm 0.0007$ & $1/2$ \\
$1/\delta_{\ReLU}$ & $0.6656 \pm 0.0001$ & $2/3$ \\
$\nu_{\rho,\mathrm{tanh}}$ & $0.4716 \pm 0.0014$ & $1/2$ \\
$\nu_{\rho,\ReLU}$ & $0.3528 \pm 0.0010$ & $1/3$ \\
\bottomrule
\end{tabular}
\end{table}

We repeat the scaling collapse for kinked activations using \relu{}.
\begin{figure*}[ht]
 \centering
 \includegraphics[width=1.0\textwidth]{\PaperRoot/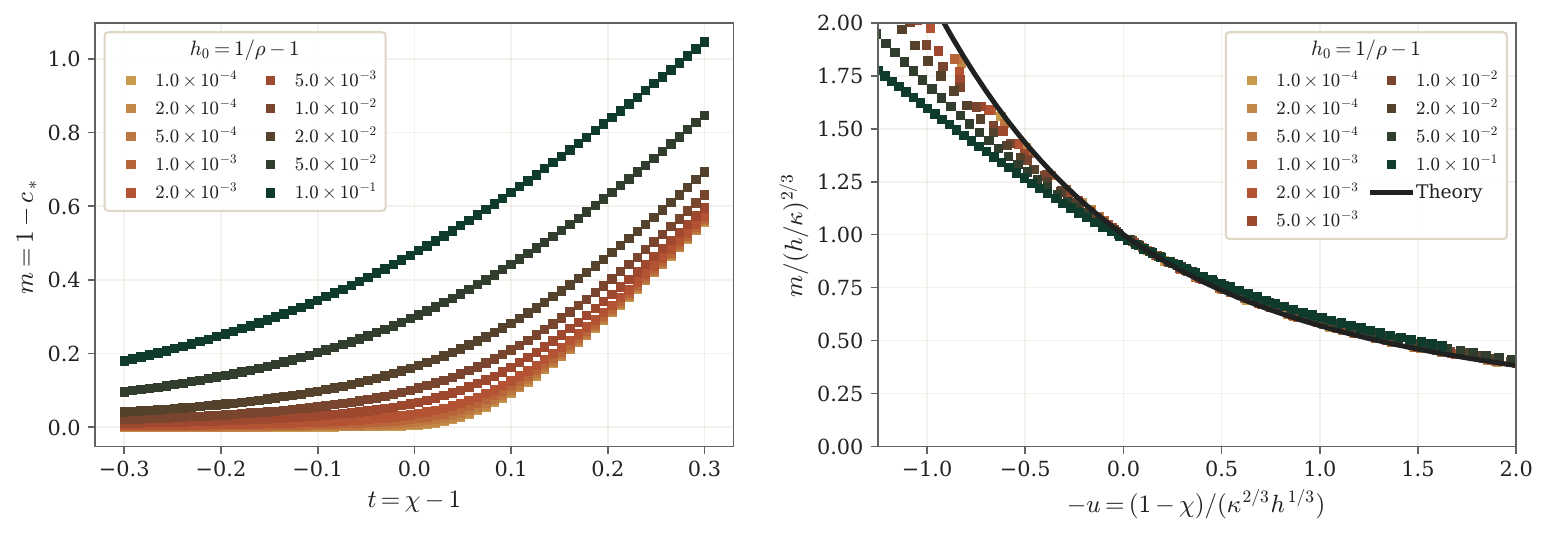}
 \caption{Kinked counterpart to Fig.~\ref{fig:uni_scaling}, using the formal \relu{} channel above. Each curve fixes $h_0=1/\rho-1$, the field at $\chi=1$. Points solve the full map and are rescaled with the actual field $h=\chi h_0$ and the critical coefficient $\kappa=2\sqrt{2}/(3\pi)$. The right panel shows the window $0\le m/(h/\kappa)^{2/3}\le2$, with horizontal coordinate $-u$ relative to Eq.~\eqref{eq:def_u}. The black curve is the leading equation of state; departures at finite field include the full map's $\chi$-dependent kink coefficient.}
 \label{fig:kinked_collapse}
\end{figure*}

\FloatBarrier

\subsection{Dropout Scheduling Experiments}

We write the mean training dropout probability as $\bar p$, distinct from the theoretical field $h$. Accuracy tables report means $\pm$ sample SEM; best accuracy is each seed's maximum recorded test accuracy, then averaged across seeds.

We test the scheduling predictions of Sec.~\ref{subsec:optimal_dropout} with MLP profiles (Fig.~\ref{fig:dropout_regimes_overfit}, Table~\ref{tab:dropout_summary_L6_run2}), controls at the same average dropout (Fig.~\ref{fig:triple_dropout}, Table~\ref{tab:triple_summary}), changes in dropout probability and width (Fig.~\ref{fig:mlp_sweeps}), ViT profiles on CIFAR-100 (Fig.~\ref{fig:dropout_regimes_transformer}, Table~\ref{tab:schedule_comparison_cifar_100}) and dropout in different Transformer components (Fig.~\ref{fig:ablations}, Table~\ref{tab:ablation_results}).

\FloatBarrier

\subsection{MLPs}
\label{app:mlps}

Concentrating dropout in early layers gives the best test performance in these MLP comparisons, at no extra computational cost. This is the placement favored by regularization reach, since an early mask can affect more of the network.

We use a setting where the model overfits, so both information propagation and regularization matter. In the effective-theory treatment of \cite{roberts2022principlesdeeplearningtheory}, deviations from Gaussian mean-field behavior \cite{neal1996bayesian,Lee2018DNNasGP} generate corrections to the correlation recursions that are suppressed by $1/N$ (with $N$ the layer width) and accumulate across depth as $L/N$. At the same time, deep information propagation suggests that successful learning is controlled by a modest multiple of the correlation length \cite{schoenholz2017deepinformationpropagation}.

We train on 5000 CIFAR-10 images with width 256, depth \(L=6\), and batch size 100. We use Gaussian weights with $\sigma_w^2=1.98$ and bias variance $\sigma_b^2=0.02$, close to the no-dropout He weight scale. This choice does not ensure stationary variance under dropout: the ReLU variance multiplier is $\sigma_w^2/(2\rho)$, which exceeds one at the tested dropout rates. The tabulated $\xi_{\rm eff}$ values come from the initialization calculation; we do not measure propagation lengths during these training runs.

\begin{figure*}[ht]
 \centering
 \includegraphics[width=0.8\textwidth]{\PaperRoot/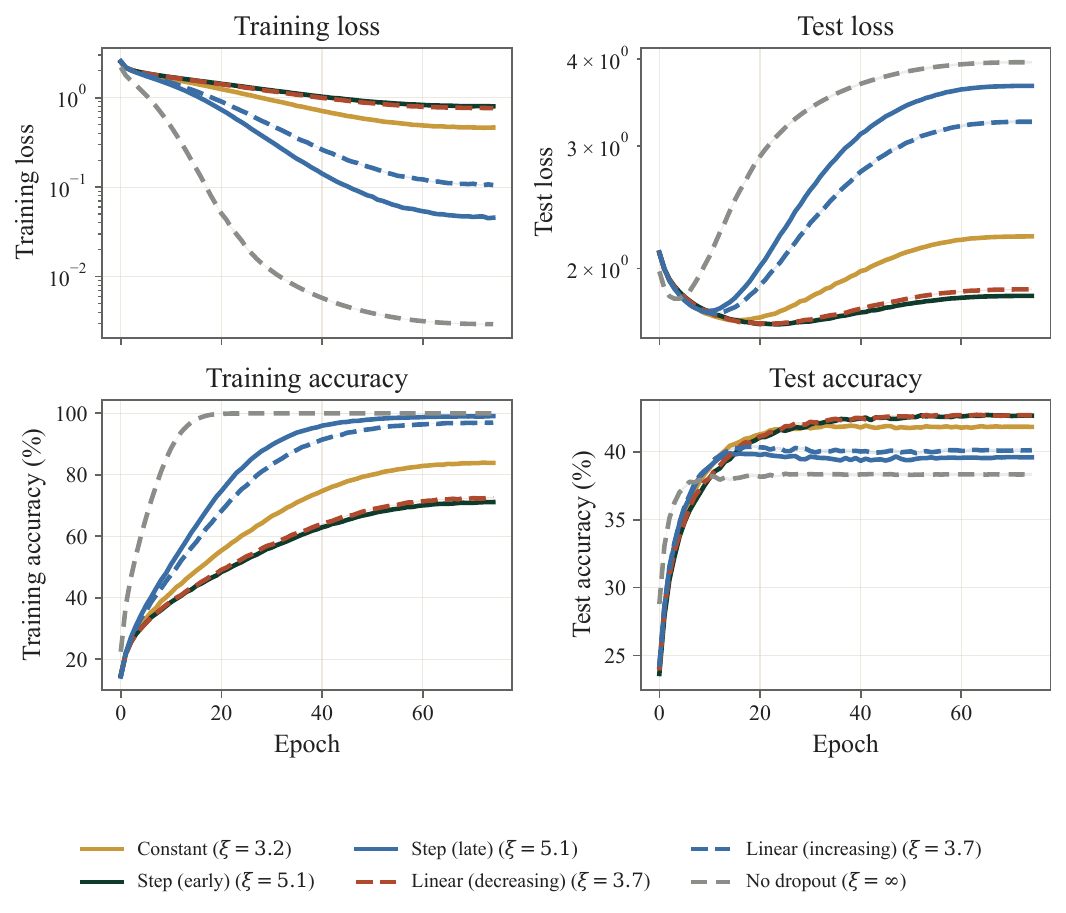}
 \caption{Finite-width MLP training and test curves at fixed mean dropout probability \(\bar p\) (with \(0 \le p_\ell \le p_{\max}\)). We compare uniform dropout, linear ramps (increasing/decreasing with depth), and step schedules that concentrate dropout in either
 the first or last half of the network, together with the no-dropout baseline.}
 \label{fig:dropout_regimes_overfit}
\end{figure*}

We compare uniform dropout, increasing and decreasing linear profiles, and step profiles concentrated in the first or last half of the network, all at the same mean dropout probability. No dropout is a separate zero-budget control (Fig.~\ref{fig:dropout_regimes_overfit}). The training code allocates raw dropout probabilities; these are proportional to the theoretical field only to leading order. Front-loaded profiles outperform both uniform and back-loaded profiles here. Early and late steps have the same reference $\xi_{\rm eff}$ but different test performance, so propagation length alone does not explain the ordering. The regularization-reach heuristic motivates the preference for early placement.

\begin{table}[!htbp]
\caption{Performance of different dropout schedules in the overfitting regime for MLPs and CIFAR-10, corresponding to Fig.~\ref{fig:dropout_regimes_overfit}.}
\label{tab:dropout_summary_L6_run2}
\centering
\begin{tabular}{lcccc}
\hline\hline
Schedule & $\xi_{\mathrm{eff}}$ & $L/\xi_{\mathrm{eff}}$ & Best Test Acc. (\%) & Final Test Acc. (\%) \\
\hline
Constant & $3.20$ & $1.87$ & $42.54 \pm 0.10$ & $41.84 \pm 0.10$ \\
Step (late) & $5.09$ & $1.18$ & $40.56 \pm 0.11$ & $39.60 \pm 0.13$ \\
Step (early) & $5.09$ & $1.18$ & $43.13 \pm 0.07$ & $42.67 \pm 0.09$ \\
Linear (increasing) & $3.73$ & $1.61$ & $41.01 \pm 0.08$ & $40.12 \pm 0.12$ \\
Linear (decreasing) & $3.73$ & $1.61$ & $43.11 \pm 0.09$ & $42.69 \pm 0.10$ \\
No dropout & $\infty$ & $0$ & $38.77 \pm 0.10$ & $38.34 \pm 0.12$ \\
\hline\hline
\end{tabular}
\end{table}

\begin{figure}[H]
 \centering
 \includegraphics[width=1.0\linewidth]{\PaperRoot/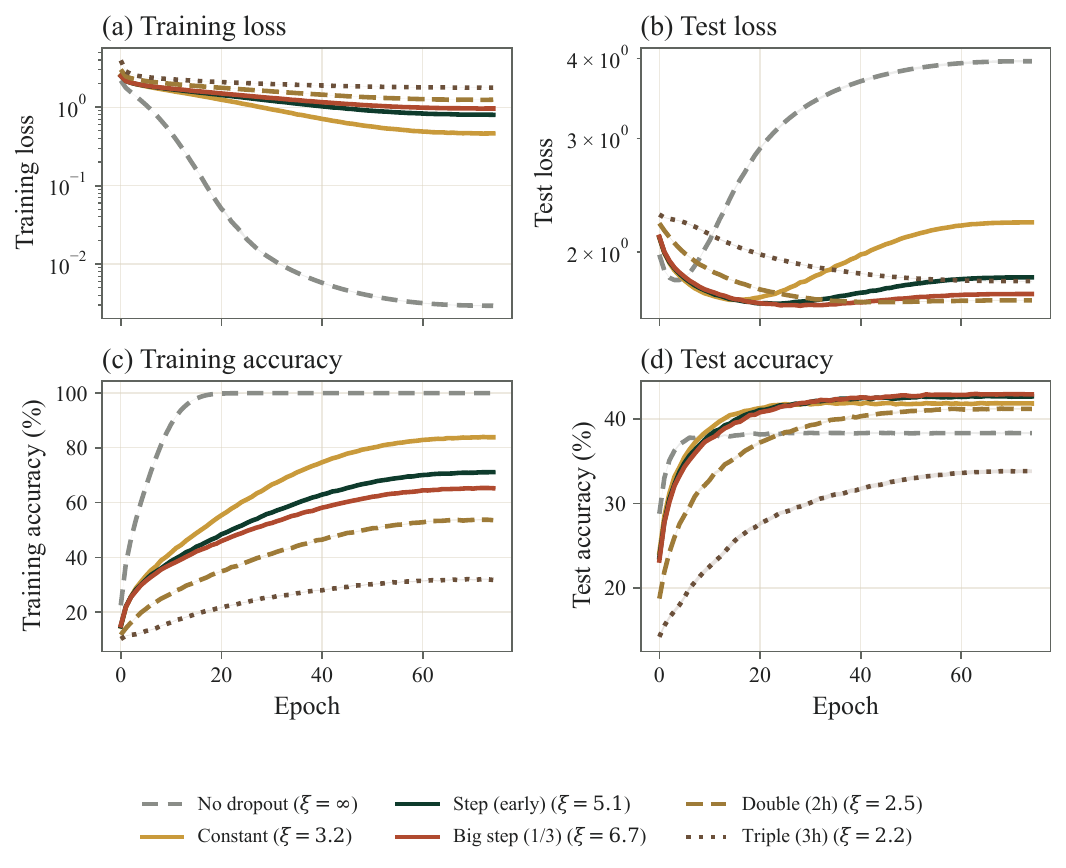}
 \caption{Front-loaded step schedules compared with constant dropout at $\bar p$, $2\bar p$ and $3\bar p$.}
 \label{fig:triple_dropout}
 \end{figure}

With six layers, uniform dropout at probability $0.1$ uses the same budget as dropout at probability $0.2$ in the first three layers. The early schedule changes where dropout acts, but also doubles its local strength, so an improvement could come from either change. We therefore also test probability $0.2$ in every layer: this matches the stronger local rate while spending twice the budget. The big-step control repeats the comparison with probability $0.3$ in the first two layers versus $0.3$ throughout (Fig.~\ref{fig:triple_dropout}, Table~\ref{tab:triple_summary}).

\clearpage

\FloatBarrier
If higher local dropout alone explains the gain, applying the same rate throughout should perform at least as well.
Applying dropout at probability $0.3$ in the first two layers gives higher accuracy than applying $0.3$ throughout, despite using one-third as much dropout (Fig.~\ref{fig:triple_dropout} and Table~\ref{tab:triple_summary}). The corresponding comparison at $0.2$ gives the same ordering. Simply increasing dropout everywhere does not reproduce the benefit of placing it early.

Early and late steps have the same reference correlation length but different accuracies, while no dropout has infinite reference length and lower accuracy. Correlation length alone therefore does not determine performance.

Higher-order terms become relevant as dropout increases. The tested probabilities up to $0.3$ probe beyond the strictly weak-dropout limit; the theory does not set a validity threshold at this value. The lower-probability comparisons are reported in the budget sweep below.

\begin{table}[ht]
\caption{Summary of dropout schedules, correlation lengths, and test performance for MLPs on CIFAR-10, corresponding to Fig.~\ref{fig:triple_dropout}.}
\label{tab:triple_summary}
\centering
\begin{tabular}{lcccc}
\hline\hline
Schedule & $\xi_{\mathrm{eff}}$ & $L/\xi_{\mathrm{eff}}$ & Best Test Acc. (\%) & Final Test Acc. (\%) \\
\hline
No dropout & $\infty$ & $0.00$ & $38.77 \pm 0.10$ & $38.34 \pm 0.12$ \\
Constant & $3.20$ & $1.87$ & $42.54 \pm 0.10$ & $41.84 \pm 0.10$ \\
Step (early) & $5.09$ & $1.18$ & $43.13 \pm 0.07$ & $42.67 \pm 0.09$ \\
Big step (1/3) & $6.67$ & $0.90$ & $43.29 \pm 0.08$ & $42.92 \pm 0.10$ \\
Double ($2\bar{p}$) & $2.54$ & $2.36$ & $41.46 \pm 0.09$ & $41.19 \pm 0.10$ \\
Triple ($3\bar{p}$) & $2.22$ & $2.70$ & $33.94 \pm 0.23$ & $33.82 \pm 0.23$ \\
\hline\hline
\end{tabular}
\end{table}

\begin{table}[h]
 \caption{Architecture and training settings for the CIFAR-10 MLPs in Fig.~\ref{fig:dropout_regimes_overfit}.}
 \label{tab:experiment_config_2}
 \centering
 \begin{tabular}{l l}
 \toprule
 \textbf{Parameter} & \textbf{Value / Description} \\
 \midrule
 \multicolumn{2}{l}{\textit{Network Architecture \& Initialization}} \\
 \midrule
 Network Width & 256 \\
 Hidden Layers & 6 \\
 Weight Variance ($\sigma_w^2$) & 1.98 \\
 Bias Variance ($\sigma_b^2$) & 0.02 \\

 \midrule
 \multicolumn{2}{l}{\textit{Training Dynamics}} \\
 \midrule
 Epochs & 75 \\
 Batch Size & 100 \\
 Optimizer & AdamW \\
 Weight Decay & $1 \times 10^{-7}$ \\
 Learning Rate & $1 \times 10^{-4}$ \\
 LR Schedule & Cosine decay to $1 \times 10^{-7}$ \\
 Mean Dropout Probability ($\bar{p}$) & 0.1 \\
 Max Dropout Probability ($p_{\max}$) & 0.2 (Step); 0.3 (Big step) \\
 Step Active Fraction ($f$) & $1/2$ (Step); $1/3$ (Big step) \\
 \midrule
 \multicolumn{2}{l}{\textit{General Setup}} \\
 \midrule
 Number of Simulations & 25 \\
 Dataset Usage & CIFAR-10 5,000 train / 5,000 test samples \\
 Hardware & GPU A100 \\
 \bottomrule
 \end{tabular}
\end{table}

\FloatBarrier

\paragraph{Sliding the dropout block.}
Figure~\ref{fig:sliding_block} uses five paired seeds (47--51) for each of four positions: layers 1--3, 2--4, 3--5 and 4--6. Each position applies dropout with probability $0.2$ after the \relu{} activation in three consecutive hidden layers, with zero dropout elsewhere. We keep the six-layer, width-256 architecture and Gaussian initialization described above, with weight variance $1.98/\mathrm{fan\mbox{-}in}$ and bias variance $0.02$. All positions use the same 5,000 training and 5,000 test images, sampled without replacement from the CIFAR-10 training and test sets with split seed 0, and the same channel normalization. We train for 75 epochs with AdamW, batch size 100, learning rate $10^{-4}$, weight decay $10^{-7}$ and cosine decay to $10^{-7}$. The figure reports the fixed final epoch; this sweep has no validation set and does not use test loss to choose an epoch or block position. Intervals are computed separately at each position as the five-seed mean $\pm t_{0.975,4}\,s/\sqrt{5}$, where $s$ is the sample standard deviation. They describe seed variation on this fixed split. These 20 fits are separate from the 25-seed schedule comparison.

\subsection{Correlations Before and After Training}
\label{app:cifar_correlations}
For Fig.~\ref{fig:cifar_correlations}, we train width-256 MLPs with twelve hidden \relu{} layers, Gaussian weights of variance $2/\mathrm{fan\mbox{-}in}$ and zero initial biases. Uniform dropout takes ten values, $p\in\{0.001,0.002,0.003,0.005,0.01,0.02,0.03,0.05,0.1,0.25\}$, each with seeds 47--51. Every run uses the same 5,000 training, 1,000 disjoint validation and 5,000 test images. We train for 35 epochs with AdamW, batch size 100, learning rate $10^{-4}$, weight decay $10^{-7}$ and cosine decay to $10^{-7}$. We save weights at epoch 35 for Fig.~\ref{fig:cifar_correlations}, and at the epoch with the lowest validation loss, taking the earlier epoch in a tie.

We select 1,024 training images once and pair each with the preceding image in a cyclic ordering.
\begin{samepage}
For each pair $(x,x')$, let $a_\ell(x)$ be its activation after dropout at hidden layer $\ell$, with $a_0(x)$ the normalized input. We estimate the uncentered correlation
\begin{equation}
 C_\ell(x,x')=
 \frac{\mathbb E_M[\langle a_\ell(x),a_\ell(x')\rangle]}
 {\sqrt{\mathbb E_M\|a_\ell(x)\|^2\,\mathbb E_M\|a_\ell(x')\|^2}},
\end{equation}
\end{samepage}
using 16 independent mask repetitions, with independent masks for the two images. We normalize each pair after averaging its moments over masks, then average over pairs. The same images and mask draws are used across rates and checkpoints; the five seeds determine the pointwise Student-$t$ intervals. Dropout remains active when probing trained weights, whereas dropout is disabled when measuring validation loss to choose the epoch. The measured mean input correlation is $0.00287$, with standard deviation $0.285$ across pairs; its small mean does not imply that individual pairs have zero correlation.

For the initialization prediction, we propagate each measured input-pair correlation through $C_{\ell+1}=(1-p)K(C_\ell)$ before averaging, where
\begin{equation}
 K(C)=\frac{\sqrt{1-C^2}+(\pi-\arccos C)C}{\pi}.
\end{equation}
The dashed levels solve $C^*=(1-p)K(C^*)$. These are infinite-depth initialization predictions; training changes the weights and biases, so we do not impose the same recursion on the trained networks.

\suppressfloats[t]
\subsection{Robustness Sweeps}
\label{app:sweeps}
\setlength{\textfloatsep}{6pt plus 1pt minus 2pt}
\setlength{\floatsep}{6pt plus 1pt minus 2pt}
\setlength{\dbltextfloatsep}{6pt plus 1pt minus 2pt}
\setlength{\dblfloatsep}{6pt plus 1pt minus 2pt}

The dropout sweeps use six-layer, width-256 MLPs. The ReLU sweep runs for 125 epochs with Gaussian weight variance $2/\mathrm{fan\mbox{-}in}$, zero initial biases and batch size 75; the 75-epoch ReLU comparisons above use weight variance $1.98/\mathrm{fan\mbox{-}in}$, bias variance $0.02$ and batch size 100. The GELU sweep runs for 75 epochs with Gaussian weight variance $2.2/\mathrm{fan\mbox{-}in}$, zero initial biases and batch size 75. The width sweep uses six hidden ReLU layers, 75 epochs, batch size 75, Gaussian weight variance $2/\mathrm{fan\mbox{-}in}$ and zero initial biases.

\begin{figure*}[t]
 \centering
 \subfigure[Sweep over mean dropout probability at width \(N=256\).]{
 \includegraphics[width=0.45\textwidth]{\PaperRoot/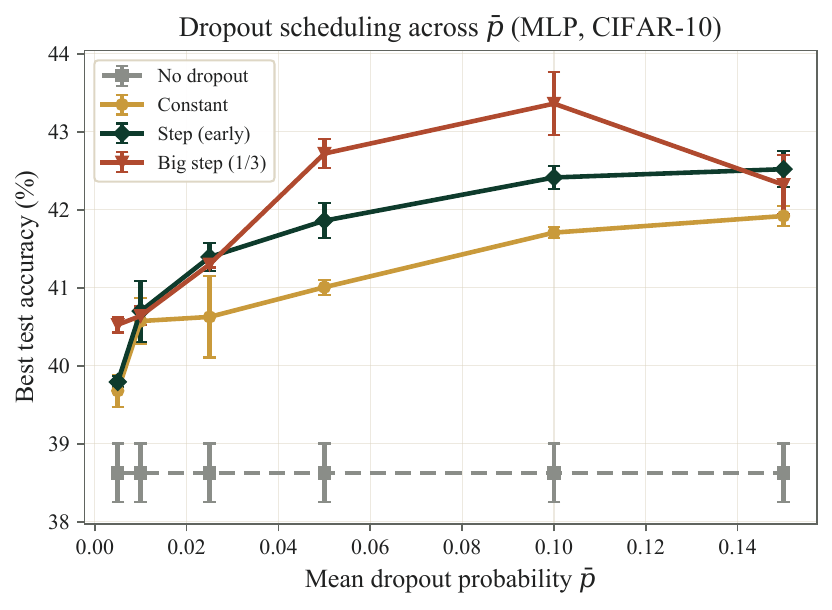}
 }
 \subfigure[Schedule advantage over constant dropout at \(\bar p=0.1\).]{
 \includegraphics[width=0.45\textwidth]{\PaperRoot/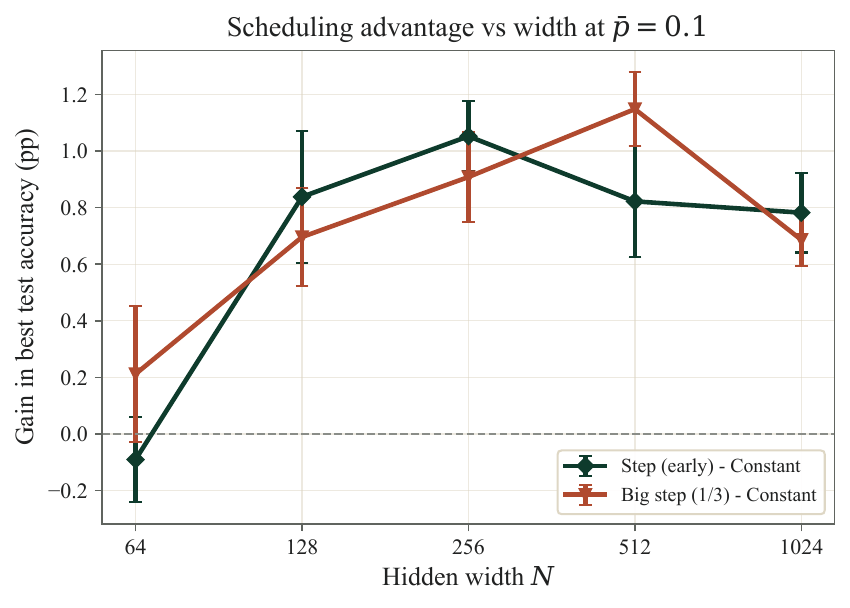}
 }
 \caption{Robustness sweeps for depth-6 near-critical \relu{} MLPs on CIFAR-10. Both early schedules improve mean best test accuracy at widths 128--1024, where the large-$N$ approximation is better justified. The differences from constant dropout are small at \(N=64\).}
 \label{fig:mlp_sweeps}
\end{figure*}

\begin{figure*}[t]
 \centering
 \includegraphics[width=0.72\textwidth]{\PaperRoot/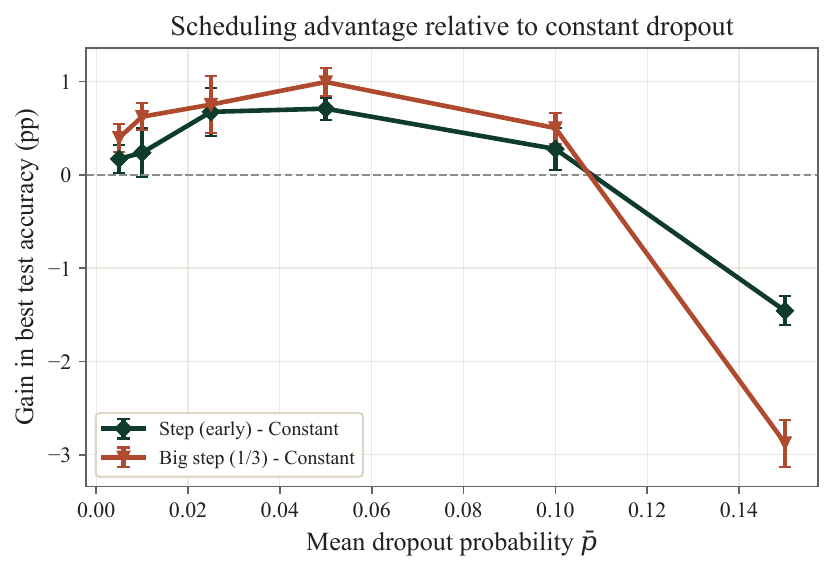}
 \caption{Smooth-activation dropout sweep for depth-6 near-critical GELU MLPs on CIFAR-10 at width \(N=256\). Step-like schedules improve over constant dropout around \(\bar p=0.1\), while the largest probability leaves the small-dropout regime where the mean-field perturbation is expected to be predictive.}
 \label{fig:gelu_h_sweep}
\end{figure*}

To test whether scheduling is tied to one budget, width, or activation, we ran CIFAR-10 MLP sweeps (Figs.~\ref{fig:mlp_sweeps} and \ref{fig:gelu_h_sweep}). In the \relu{} dropout sweep, Step/Big step use large local dropout probabilities \(0.30/0.45\) at \(\bar p=0.15\). Both early schedules improve mean best test accuracy at widths 128--1024, where the large-$N$ approximation is better justified; their differences from uniform are small at \(N=64\). The GELU MLP sweep gives the same ordering in maximum test accuracy at \(\bar p=0.1\)--constant \(41.96\pm0.16\), Step \(42.24\pm0.15\), Big step \(42.46\pm0.12\)--and Big step falls to \(38.97\pm0.23\) at \(\bar p=0.15\), consistent with the small-dropout approximation becoming less accurate. The final-epoch gains summarized in Table~\ref{tab:loss_improvements} use the last recorded epoch, giving \(+1.56\) pp for the \relu{} \(\bar p=0.1\) comparison after extension to ten pairs and \(+0.62\) pp for the unchanged ten-pair GELU comparison. The full sweeps shown here retain their original runs.

\FloatBarrier

\subsection{Transformers on CIFAR-100}
\begin{figure*}[h]
 \centering
 \includegraphics[width=0.95\textwidth]{\PaperRoot/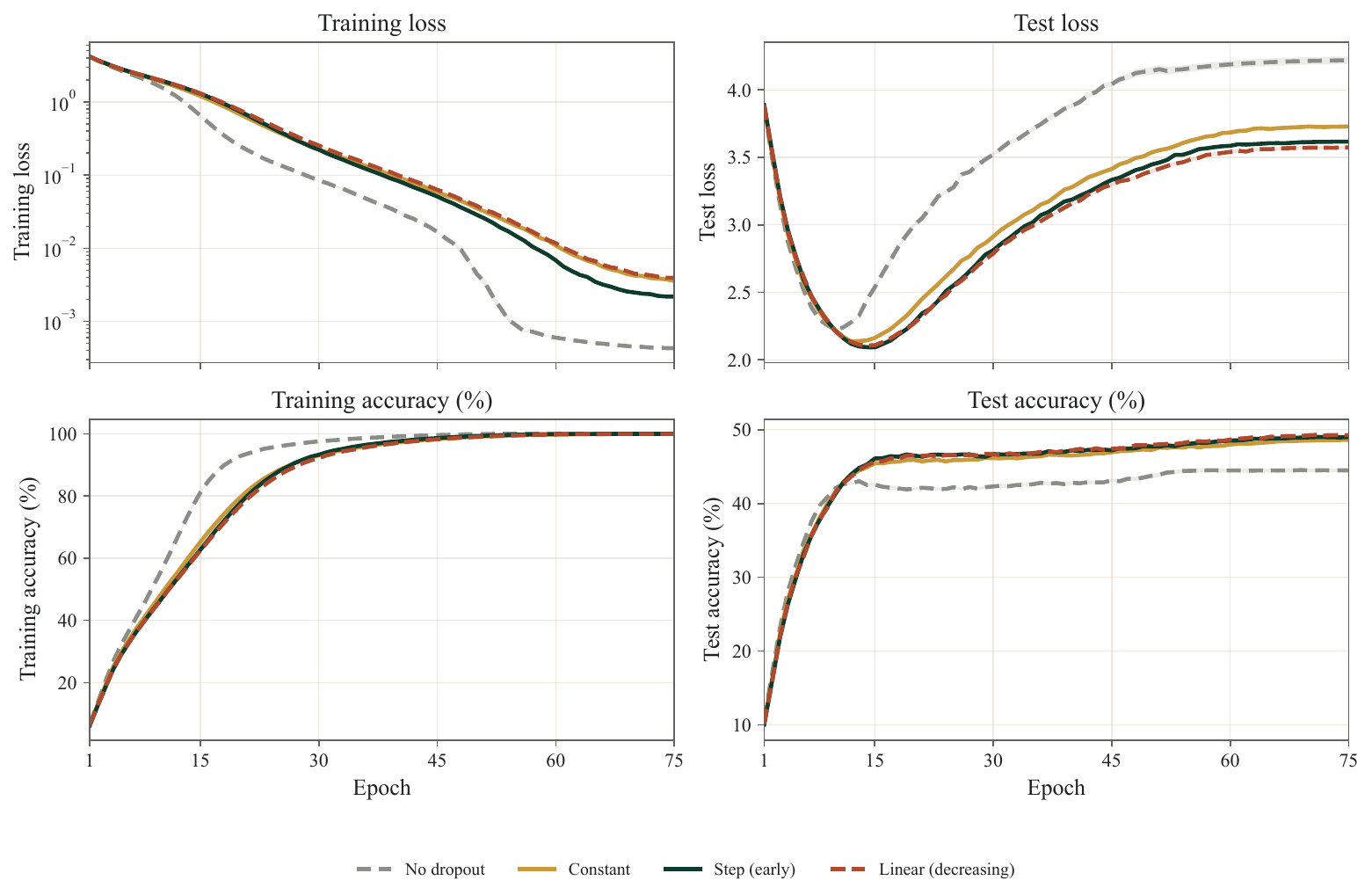}
 \caption{Vision Transformer training and test curves on CIFAR-100 over all 75 epochs. Uniform, linearly decreasing and early-step dropout share the same mean probability \(\bar p\); no dropout is shown as a control. Lines give means over ten seeds; shading gives SEM.}
 \label{fig:dropout_regimes_transformer}
\end{figure*}

Table~\ref{tab:vit_config} gives the architecture and training settings for the CIFAR-100 Vision Transformer \cite{dosovitskiy2021vit,krizhevsky2009cifar}, with learning curves in Fig.~\ref{fig:dropout_regimes_transformer} and test accuracy in Table~\ref{tab:schedule_comparison_cifar_100}.

\begin{table}[h]
 \caption{Reported settings for the CIFAR-100 Vision Transformer. The saved metrics contain ten 75-epoch runs per profile. Their accuracy increments require 50,000 training and 10,000 test images under the archived evaluator, supporting use of the full dataset. Architecture and optimizer settings come from the code and reported recipe; the metrics do not retain a configuration that verifies those settings for these runs.}
 \label{tab:vit_config}
 \centering
 \footnotesize
 \setlength{\tabcolsep}{3pt}
 \renewcommand{\arraystretch}{0.9}
 \begin{tabular}{l l}
 \toprule
 \textbf{Parameter} & \textbf{Value / Description} \\
 \midrule
 \multicolumn{2}{l}{\textit{ViT Architecture}} \\
 \midrule
 Embedding Dimension & 128 \\
 Depth & 10 \\
 Number of Heads & 16 \\
 MLP Ratio & 4.0 \\
 Patch Size & 4 \\
 \midrule
 \multicolumn{2}{l}{\textit{Training Dynamics}} \\
 \midrule
 Epochs & 75 \\
 Batch Size & 250 \\
 Learning Rate & $5 \times 10^{-4}$ \\
 LR Minimum & $5 \times 10^{-6}$ \\
 Weight Decay & 0.00 \\
 Mean Dropout Probability ($\bar{p}$) & 0.1 \\
 Max Dropout Probability ($p_{\max}$) & 0.2 \\
 Schedules & None, Constant, Step (early), Linear (decreasing) \\
 \midrule
 \multicolumn{2}{l}{\textit{General Setup}} \\
 \midrule
 Number of Simulations & 10 \\
 Dataset Usage & Full CIFAR-100 \\
 Evaluation Frequency & Every 1 Epoch \\
 Hardware & GPU A100 \\
 \bottomrule
 \end{tabular}
\end{table}

On CIFAR-100, linearly decreasing dropout raises mean final accuracy from $48.61\%$ under uniform dropout to $49.27\%$, while early-step dropout reaches $49.01\%$ (Table~\ref{tab:schedule_comparison_cifar_100}, Fig.~\ref{fig:dropout_regimes_transformer}). Errors and curve bands show SEM. These comparisons test the schedules empirically on Transformers; the propagation calculation covers MLPs.

\begin{table}[H]
 \caption{Test accuracy across different dropout schedules for ViT architecture on CIFAR-100, corresponding to Fig.~\ref{fig:dropout_regimes_transformer}.}
 \label{tab:schedule_comparison_cifar_100}
 \centering
 \footnotesize
 \begin{tabular}{l c c}
 \toprule
 \textbf{Schedule} & \textbf{Best Test Acc} & \textbf{Final Test Acc} \\
 \midrule
 No dropout & $44.61 \pm 0.25$ & $44.50 \pm 0.25$ \\
 Constant & $48.69 \pm 0.18$ & $48.61 \pm 0.20$ \\
 Step (early) & $49.10 \pm 0.11$ & $49.01 \pm 0.12$ \\
 Linear (decreasing) & $49.38 \pm 0.14$ & $49.27 \pm 0.14$ \\
 \bottomrule
 \end{tabular}
\end{table}

\FloatBarrier

\subsection{Transformer Ablations}
\begin{figure}[H]
 \centering
 \includegraphics[width=1\linewidth]{\PaperRoot/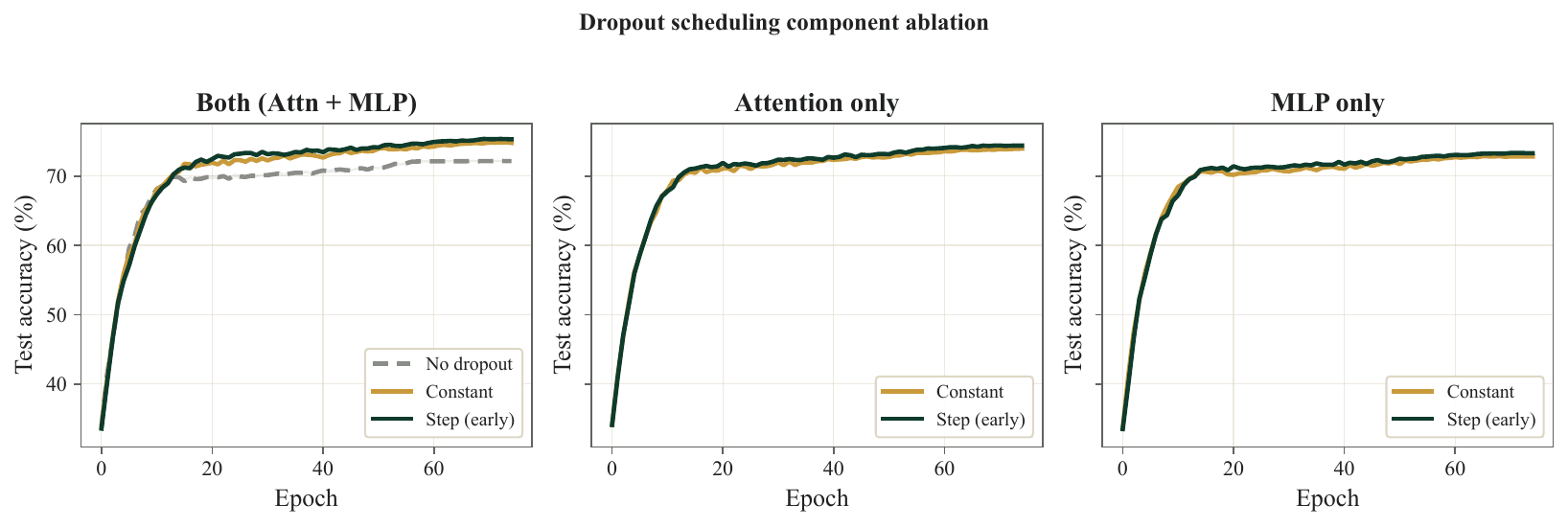}
 \caption{Dropout applied to the attention block, MLP block or both, comparing no dropout, constant dropout and step-like dropout over five seeds. The five additional seeds with dropout in both blocks are not included here.}
 \label{fig:ablations}
\end{figure}

We apply dropout to the attention block, the MLP block or both, comparing step-like and constant schedules against no dropout to see which components benefit from early placement. The original ablation uses the full CIFAR-10 dataset, five seeds, and a 12-block ViT with embedding dimension 128, 16 attention heads, MLP ratio 4, and patch size 4. Training runs for 75 epochs on an A100 with batch size 1000, learning rate $5\times10^{-4}$ decaying to $5\times10^{-6}$, and zero weight decay. Regularized profiles use $\bar p=0.1$ and $p_{\max}=0.2$; evaluation is recorded every epoch. Figure~\ref{fig:ablations} and Table~\ref{tab:ablation_results} show these five-seed comparisons. The main table adds five fresh pairs for uniform and early step with dropout in both blocks.

\begin{table}[H]
 \caption{Dropout in the MLP block, attention block or both on CIFAR-10, comparing constant and step-like schedules. Means $\pm$ sample SEM over five seeds. Best accuracy averages each seed's maximum recorded test accuracy; final accuracy uses the last epoch. The early-step mean exceeds the uniform mean in all three configurations, without establishing significance for each comparison.}
 \label{tab:ablation_results}
 \centering
 \begin{tabular}{l c c}
 \toprule
 \textbf{Configuration} & \textbf{Best Test Acc} & \textbf{Final Test Acc} \\
 \midrule
 No dropout & $72.22 \pm 0.16$ & $72.15 \pm 0.16$ \\
 \midrule
 MLP only (Constant) & $72.96 \pm 0.13$ & $72.82 \pm 0.14$ \\
 MLP only (Step early) & $73.38 \pm 0.28$ & $73.30 \pm 0.30$ \\
 \midrule
 Attn only (Constant) & $74.00 \pm 0.10$ & $73.95 \pm 0.11$ \\
 Attn only (Step early) & $74.48 \pm 0.10$ & $74.38 \pm 0.10$ \\
 \midrule
 Both (Constant) & $74.92 \pm 0.11$ & $74.78 \pm 0.11$ \\
 Both (Step early) & $75.47 \pm 0.14$ & $75.30 \pm 0.13$ \\
 \bottomrule
 \end{tabular}
\end{table}

\FloatBarrier

\FloatBarrier
\raggedbottom
\section{Extended Experiments Across Datasets}
\label{app:extended_experiments}
We compare dropout placement across datasets, architectures and training-set sizes, using five paired seeds for each profile and keeping the data split and training procedure fixed within each comparison. Table~\ref{tab:frontloaded_results} combines all available seeds for uniform and the selected front-loaded profiles, reporting the seed count separately for each evaluation. The figures below retain five seeds per profile; Table~\ref{tab:fi2010_standalone_comparison} reports the separate FI-2010 study. Section~\ref{app:benchmark_methods} gives the architectures, data splits, optimizer settings, search grids and seeds, and identifies settings that could not be recovered.

\paragraph{Choosing profiles and measuring performance.}
We tune learning rate and mean dropout separately for each profile using validation data. For each completed five-seed comparison, we average the lowest validation CE reached by each seed, then choose the nonuniform profile with the lowest average. Test CE and accuracy are measured at the epoch with the lowest validation loss in each run; we keep the chosen profile even where it loses to uniform. Of these five-seed experiments, only Tiny ImageNet 80k saved test CE at the final epoch. Results are means $\pm$ SEM and curve bands show SEM over five seeds unless stated otherwise; paired CE differences subtract uniform from the named profile. No-dropout controls have their own tuned learning rates. In the separate FI-2010 study, we instead choose the profile retrospectively from mean test CE.

\paragraph{Curves and profiles.}
The benchmark figures show training and validation curves for the first five seeds; the separate CIFAR-10 pilot uses three. Where both models appear, the MLP is above the Transformer; columns show training CE, validation CE, training accuracy, and validation accuracy. Training CE has a logarithmic axis. Training metrics average batches with dropout enabled, while validation evaluates each epoch's final weights with dropout disabled. Step early fills the earliest layers at probability $0.20$, with a partially filled last active layer when needed. Big step uses $3\bar p$ in the first third; linear profiles run between $0$ and $2\bar p$. Thus profiles can differ in both mean and peak dropout. Transformer profiles are empirical extensions; the MLP reference-field diagnostic is only a proxy.

\subsection{Data and training}
\label{app:benchmark_methods}
\label{app:benchmark_protocols}
Tables~\ref{tab:benchmark_protocols} and~\ref{tab:benchmark_search} give the data and training settings. The repository contains the preprocessing and training code, together with the recovered settings and data splits. Missing settings are identified below.

\begin{table}[ht]
\centering
\caption{Data and training settings. M and T denote MLP and Transformer. $E$ is the epoch count and WD is weight decay. All rows use batch size 128. Counts are training/validation/test examples. The standalone FI-2010 counts follow the available segment manifest; its match to the original trial manifest is unverified.}
\label{tab:benchmark_protocols}
\small
\setlength{\tabcolsep}{4pt}
\renewcommand{\arraystretch}{1.2}
\begin{tabular}{@{}p{0.18\textwidth}p{0.37\textwidth}p{0.20\textwidth}p{0.18\textwidth}@{}}
\toprule
Dataset & Input and split & Counts & Models; $E$; WD \\
\midrule
Speech Commands & 35 classes; $64\!\times\!64$ log-mel image; balanced random clips from official training subset & 20,020/5,005/10,010 & M,T; 50; $0$ \\
Tiny ImageNet 20k & 200 classes; $3\!\times\!64\!\times\!64$ RGB; balanced subsets of official training images & 20,000/5,000/10,000 & M,T; 75; $0$ \\
Tiny ImageNet 80k & Same image pool; separately drawn balanced split & 80,000/10,000/10,000 & M,T; 75; $0$ \\
FI-2010 benchmark & 3 classes; $100\!\times\!40$ book window; one stock, forward split with 100-window gaps & 40,000/10,000/20,000 & M; 50; $0$ \\
FI-2010 standalone & $128\!\times\!40$ windows within each stock/day; train day 6, validation day 7, test days 8--10 & 38,467/36,661/137,532 & M; 100; $0$ \\
\bottomrule
\end{tabular}
\end{table}

\paragraph{Preparing the inputs.}
Continuous inputs use coordinate-wise training-set means and population standard deviations; a standard deviation below $10^{-8}$ is replaced by one. There is no data augmentation. Speech waveforms use the first channel, resample to 16\,kHz when needed, and truncate or zero-pad to one second. The mel transform uses FFT size 1024, hop 250, 64 mel bands, and an 80\,dB range; the code retains the first 64 frames. A reconstruction of the random clip split finds that 4,979 of 5,005 validation clips and 9,967 of 10,010 test clips share a speaker with training. These results therefore measure performance on held-out clips, mostly from familiar speakers. The 80\,dB cutoff was applied across each 256-clip preprocessing batch. Tiny ImageNet uses RGB pixels divided by 255 and sorted class/file order. The 20k experiment computes normalization statistics in float32; the 80k experiment combines float64 moments in 64\,MiB chunks before casting the statistics to float32. Their holdout sets differ. Random class-balanced splits use NumPy's \texttt{default\_rng} with seed 20260812, sampling without replacement within each class.

The benchmark FI-2010 cache reads the first 40 feature rows of the no-auction Z-score release's ninth cumulative training fold, within stock ordinal 5 only; labels use zero-based row 148, the fifth supplied horizon, at the end of the window. Its sequential window-index ranges are $[0,40000)$, $[40100,50100)$, and $[50200,70200)$. The standalone study instead uses decimal-precision daily segments, all five stock ordinals, and label row 144 (the first horizon). Its 128-row windows never cross stock/day boundaries, and the last ten representations of each segment are purged. Its feature normalization uses day-6 rows only, excluding these tails, with float64 moments cast to float32.

\paragraph{Architectures.}
Benchmark MLPs have 12 hidden affine--ReLU--dropout blocks of width 256 and a separate affine classifier. Every affine layer, including the readout, has independent Gaussian weights of variance $1.98/\mathrm{fan\_in}$ and biases of variance $0.02$. Transformers have 12 pre-LayerNorm residual blocks, width 256, eight attention heads, and a ReLU feedforward sublayer of width 1024. The layer's dropout probability is applied separately to the attention output and feedforward output before residual addition; attention weights have no dropout. Learned absolute positions and a learned class token precede a final LayerNorm and class-token classifier. Speech and Tiny ImageNet use $8\!\times\!8$ patches (64 patches). Linear weights and position/class embeddings use truncated-normal initialization with standard deviation $0.02$; linear biases are zero. Patch convolutions retain the library's default initialization. Standalone MLPs use the same Gaussian affine initialization with depths $\{6,12\}$ and widths $\{256,512\}$.

\paragraph{Training and evaluation.}
Benchmarks use AdamW, zero weight decay, unweighted cross-entropy, and no warmup. At epoch $e\in\{0,\ldots,E-1\}$ the learning rate is
\[
\eta_e=\eta_0\left[f+\frac{1-f}{2}\left(1+\cos\frac{\pi e}{E-1}\right)\right],
\qquad f=10^{-3}.
\]
Transformers clip the gradient norm at one; benchmark MLPs do not clip. Standalone MLPs use Adam, norm clipping at one, and the same cosine formula with $f=10^{-2}$. Both optimizers retain $(\beta_1,\beta_2)=(0.9,0.999)$ and $\epsilon=10^{-8}$. Training shuffles examples and retains the last partial batch. Benchmark CUDA training uses autocast (bfloat16 when supported, otherwise float16); standalone training uses float32 with deterministic algorithms and TF32 disabled.

We evaluate each run on test data using the weights from the epoch with the lowest validation loss, taking the earlier epoch in a tie. The first five seeds include final-epoch test results only for Tiny ImageNet 80k; the five additional seeds include them for every dataset. Final results in Table~\ref{tab:frontloaded_results} therefore use ten pairs for Tiny ImageNet 80k and five elsewhere. Test performance was not recorded at every epoch.

\par\medskip
\noindent\begin{minipage}{\textwidth}
\begingroup
\centering
\captionof{table}{Hyperparameter searches and seeds. The grids and selection rules come from the training code, with settings for the benchmark runs checked against saved configuration hashes. Additional seeds keep each profile's settings unchanged.}
\label{tab:benchmark_search}
\small
\setlength{\tabcolsep}{4pt}
\renewcommand{\arraystretch}{1.2}
\begin{tabular}{@{}p{0.19\textwidth}p{0.49\textwidth}p{0.26\textwidth}@{}}
\toprule
Experiment & Search and chosen settings & Seeds \\
\midrule
Zero weight decay & Per-profile LR search at $\bar p=0.10$: MLP $\{3\!\times\!10^{-5},10^{-4},3\!\times\!10^{-4},10^{-3},3\!\times\!10^{-3}\}$; Transformer replaces the first value by $5\!\times\!10^{-5}$. Then search $\bar p\in\{0.05,0.10,0.15,0.20\}$ at that profile's selected LR. No dropout has its own LR search. & LR: $0$; budget: $0,1,2$; all profiles: $100$--$104$; extension: $105$--$109$ \\
FI-2010 standalone & Uniform-only 50-epoch LR screen over $\{3\!\times\!10^{-5},10^{-4},3\!\times\!10^{-4},10^{-3}\}$; keep one LR per depth and width across seven profiles, with $\bar p=0.10$ except no dropout. & LR: $0,1,2$; 100-epoch confirmation: $100$--$104$ \\
\bottomrule
\end{tabular}
\par
\endgroup

\paragraph{Choosing the settings.}
For each setting, we average the lowest validation cross-entropy reached by each seed and choose the setting with the lowest average. Ties favor the lower learning rate, then the lower budget in the budget sweep. These hyperparameters are chosen before running the confirmation seeds. For Table~\ref{tab:frontloaded_results}, we choose the front-loaded profile by the same average over the five confirmation seeds, considering only profiles with all five paired runs complete. This profile choice is retrospective. Additional seeds keep that profile and its hyperparameters fixed, and the same profile is used for test results at the epoch with the lowest validation loss and at the final epoch.

\paragraph{Uncertainty across seeds.}
We pair profiles by seed, using separate random streams for initialization, minibatch order and dropout. The nominal 95\% intervals use Fieller's method for relative loss reductions and Student-$t$ intervals for accuracy gains. They measure variation across seeds for the chosen split and settings, without accounting for the earlier selection or multiple comparisons. Both use $n-1$ degrees of freedom for $n$ independent seed pairs. Fieller intervals assume approximately bivariate-normal paired losses; Student-$t$ intervals assume approximately normal paired accuracy differences.

\paragraph{Missing settings.}
The complete search results could not be recovered. For standalone FI-2010, the selected learning rates and hashes of the original runs, source and data are also missing. The released code repeats the documented search and training procedure, with missing values marked in the repository. Source hashes identify the recovered benchmark code; results need not be bitwise identical across library versions or devices.

\paragraph{Additional seeds.}
Seven benchmark comparisons add seeds $105$--$109$ to the original $100$--$104$, giving ten paired test measurements at the epoch with the lowest validation loss for each row in Table~\ref{tab:frontloaded_results}. The extension also brings the original CIFAR-10 ReLU comparison from three to ten pairs (new seeds $45$--$51$) and the both-block ViT comparison from five to ten (new seeds $47$--$51$). These 94 additional runs use the same profiles, hyperparameters and data splits, without a new search. We checked the benchmark settings and hashes of the full cached data splits before combining results. The additional runs used H100 GPUs, while the CIFAR experiments reported A100s. The ReLU implementation also required a change to tensor storage before flattening, preserving the order of the features. The repository contains the settings and measurements, with the initial and additional seeds listed separately.
\end{minipage}
\par\medskip

\FloatBarrier
\noindent\begin{minipage}{\textwidth}
\subsection{Speech Commands}
We train the MLP and Transformer for 50 epochs without weight decay, using the same 20,020 training, 5,005 validation and 10,010 test clips. Clips are drawn randomly from the official training subset, so most test speakers also appear in training.

Across the five seeds shown here, early-step dropout reduces the MLP's test loss by $0.136\pm0.007$, while big-step dropout reduces the Transformer's by $0.036\pm0.037$, measured at the epoch with the lowest validation loss. Errors are standard errors of the paired differences from uniform.

The reductions in minimum and final validation loss are $13.23\%$ and $13.47\%$ for the MLP, and $3.33\%$ and $6.49\%$ for the Transformer. These figures use the five seeds plotted here; the main results table includes the additional seeds (Table~\ref{tab:frontloaded_results}).

\par\medskip
\begingroup
\centering
\includegraphics[width=\textwidth]{\PaperRoot/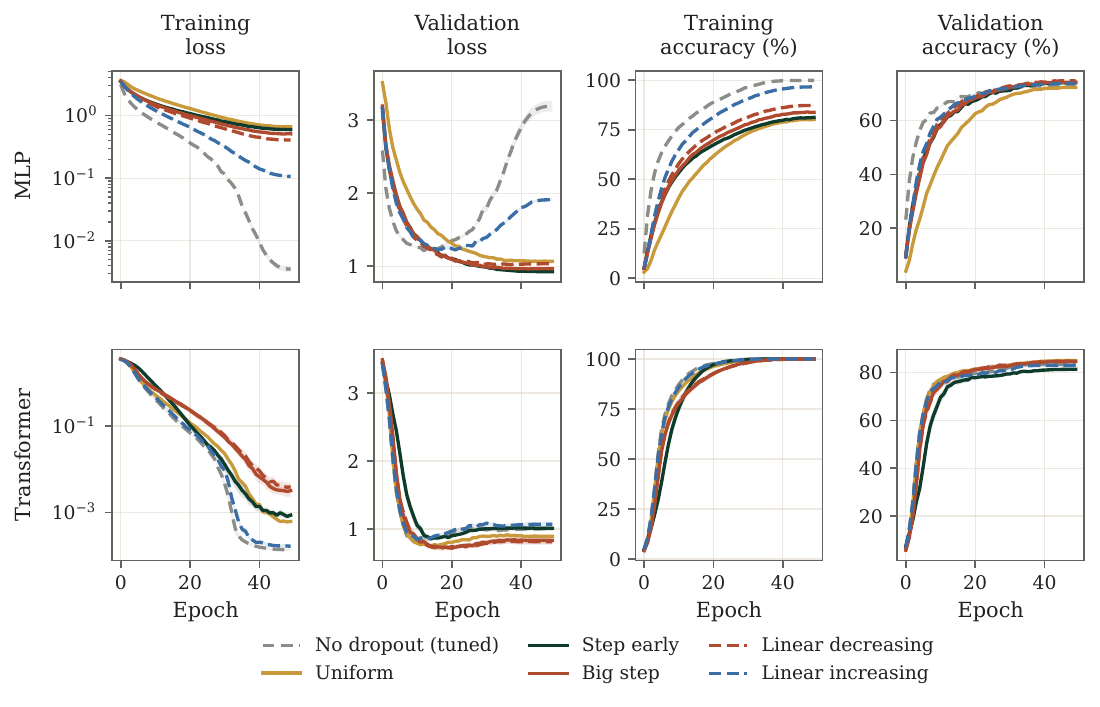}
\par\medskip
\begingroup\small
\setlength{\tabcolsep}{3pt}
\begin{tabular}{@{}lrrrr@{\hspace{10pt}}rrrr@{}}
\toprule
Profile & \multicolumn{4}{c}{MLP} & \multicolumn{4}{c}{Transformer} \\
\cmidrule(lr){2-5}\cmidrule(lr){6-9}
 & $\bar p$ & LR & Test CE & Acc. (\%) & $\bar p$ & LR & Test CE & Acc. (\%) \\
\midrule
No dropout & 0.00 & 1e-03 & \(1.203\pm0.011\) & \(67.16\pm0.35\) & 0.00 & 3e-04 & \(0.807\pm0.018\) & \(77.65\pm0.65\) \\
Uniform & 0.10 & 1e-03 & \(1.073\pm0.006\) & \(72.23\pm0.12\) & 0.15 & 3e-04 & \(0.743\pm0.017\) & \(79.54\pm0.61\) \\
Step early & 0.05 & 1e-03 & \(0.937\pm0.003\) & \(74.06\pm0.13\) & 0.05 & 1e-04 & \(0.852\pm0.027\) & \(77.85\pm0.64\) \\
Big step & 0.05 & 1e-03 & \(0.950\pm0.004\) & \(74.07\pm0.11\) & 0.15 & 3e-04 & \(0.707\pm0.036\) & \(81.07\pm0.89\) \\
Linear decreasing & 0.05 & 1e-03 & \(1.002\pm0.005\) & \(73.87\pm0.45\) & 0.20 & 3e-04 & \(0.711\pm0.033\) & \(81.27\pm1.05\) \\
Linear increasing & 0.05 & 1e-03 & \(1.202\pm0.009\) & \(69.40\pm0.54\) & 0.10 & 3e-04 & \(0.832\pm0.022\) & \(77.67\pm0.67\) \\
\bottomrule
\end{tabular}
\endgroup
\captionof{figure}{Speech Commands: learning curves and test results for five seeds on randomly held-out clips, with speakers shared across splits. Both no-dropout controls have separately tuned learning rates. The table reports test results at the epoch with the lowest validation loss, while the curves show training and validation performance.}
\label{fig:speech-commands}
\endgroup\par
\end{minipage}\par\bigskip

\noindent\begin{minipage}{\textwidth}
\subsection{Tiny ImageNet: 20,000 training examples}
Both models train for 75 epochs with zero weight decay on a fixed 20,000/5,000/10,000 training/validation/test split. Step early, chosen using validation loss, has paired differences in test CE from uniform of \(-0.094\pm0.011\) for the MLP and \(-0.086\pm0.013\) for the Transformer.

\par\medskip
\begingroup
\centering
\includegraphics[width=\textwidth]{\PaperRoot/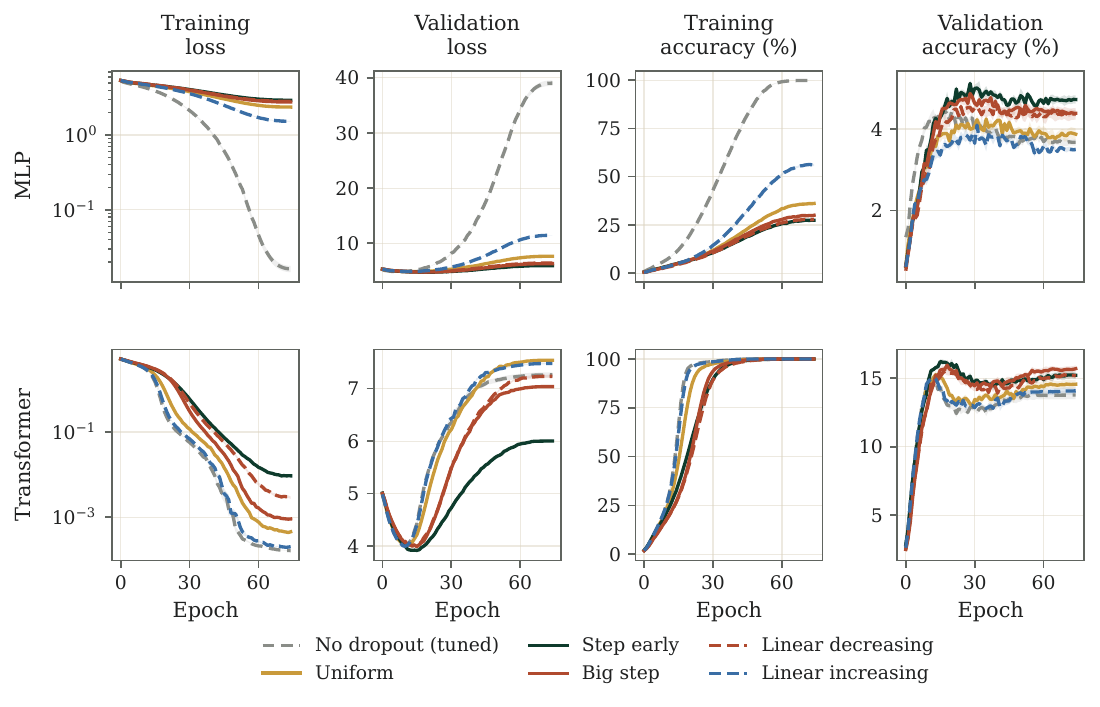}
\par\medskip
\begingroup\small
\setlength{\tabcolsep}{3pt}
\begin{tabular}{@{}lrrrr@{\hspace{10pt}}rrrr@{}}
\toprule
Profile & \multicolumn{4}{c}{MLP} & \multicolumn{4}{c}{Transformer} \\
\cmidrule(lr){2-5}\cmidrule(lr){6-9}
 & $\bar p$ & LR & Test CE & Acc. (\%) & $\bar p$ & LR & Test CE & Acc. (\%) \\
\midrule
No dropout & 0.00 & 1e-03 & \(4.856\pm0.005\) & \(3.88\pm0.05\) & 0.00 & 3e-04 & \(3.981\pm0.012\) & \(14.92\pm0.21\) \\
Uniform & 0.05 & 1e-03 & \(4.868\pm0.006\) & \(3.45\pm0.06\) & 0.10 & 3e-04 & \(3.976\pm0.007\) & \(14.63\pm0.17\) \\
Step early & 0.05 & 3e-04 & \(4.774\pm0.006\) & \(4.84\pm0.04\) & 0.15 & 1e-04 & \(3.890\pm0.008\) & \(16.26\pm0.10\) \\
Big step & 0.05 & 3e-04 & \(4.787\pm0.003\) & \(4.61\pm0.10\) & 0.15 & 3e-04 & \(3.960\pm0.012\) & \(15.45\pm0.29\) \\
Linear decreasing & 0.05 & 3e-04 & \(4.823\pm0.004\) & \(4.38\pm0.10\) & 0.20 & 3e-04 & \(3.980\pm0.011\) & \(14.71\pm0.22\) \\
Linear increasing & 0.05 & 1e-03 & \(4.898\pm0.007\) & \(3.36\pm0.09\) & 0.05 & 3e-04 & \(3.981\pm0.017\) & \(14.73\pm0.30\) \\
\bottomrule
\end{tabular}
\endgroup
\captionof{figure}{Tiny ImageNet 20k: learning curves and test results for five seeds, including all six profiles and both no-dropout controls.}
\label{fig:tiny-imagenet-20k}
\endgroup\par
\end{minipage}\par\bigskip

\noindent\begin{minipage}{\textwidth}
\subsection{Tiny ImageNet: 80,000 training examples}
This separate balanced split has 80,000/10,000/10,000 training/validation/test examples; both models train for 75 epochs with zero weight decay. Every listed configuration uses learning rate $3\times10^{-4}$. The selected profiles' paired differences in test CE from uniform are \(-0.058\pm0.005\) for the MLP (step early) and \(-0.031\pm0.015\) for the Transformer (linear decreasing). Test CE at the final epoch is available for every profile in this comparison.

\par\medskip
\begingroup
\centering
\includegraphics[width=\textwidth]{\PaperRoot/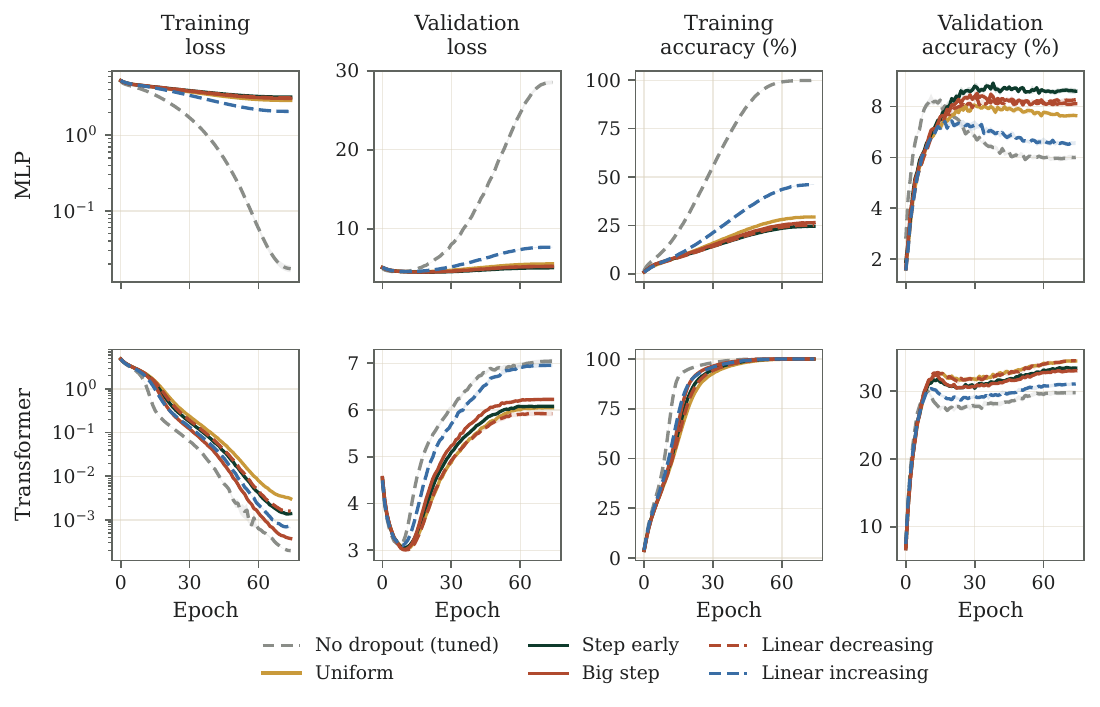}
\par\medskip
\begingroup\small
\setlength{\tabcolsep}{3pt}
\begin{tabular}{@{}lrrrr@{\hspace{10pt}}rrrr@{}}
\toprule
Profile & \multicolumn{4}{c}{MLP} & \multicolumn{4}{c}{Transformer} \\
\cmidrule(lr){2-5}\cmidrule(lr){6-9}
 & $\bar p$ & Test CE & Acc. (\%) & Final CE & $\bar p$ & Test CE & Acc. (\%) & Final CE \\
\midrule
No dropout & 0.00 & \(4.561\pm0.004\) & \(7.64\pm0.15\) & \(28.876\pm0.230\) & 0.00 & \(3.125\pm0.025\) & \(29.57\pm0.44\) & \(7.126\pm0.068\) \\
Uniform & 0.05 & \(4.575\pm0.002\) & \(7.09\pm0.14\) & \(5.627\pm0.004\) & 0.20 & \(3.035\pm0.014\) & \(31.49\pm0.17\) & \(6.104\pm0.043\) \\
Step early & 0.05 & \(4.518\pm0.004\) & \(7.95\pm0.10\) & \(5.130\pm0.009\) & 0.15 & \(3.053\pm0.015\) & \(31.22\pm0.21\) & \(6.164\pm0.025\) \\
Big step & 0.05 & \(4.541\pm0.003\) & \(7.53\pm0.12\) & \(5.321\pm0.007\) & 0.10 & \(3.026\pm0.013\) & \(31.35\pm0.38\) & \(6.296\pm0.034\) \\
Linear decreasing & 0.05 & \(4.563\pm0.003\) & \(7.19\pm0.15\) & \(5.286\pm0.013\) & 0.15 & \(3.004\pm0.016\) & \(31.78\pm0.23\) & \(6.002\pm0.042\) \\
Linear increasing & 0.05 & \(4.608\pm0.005\) & \(6.75\pm0.12\) & \(7.764\pm0.023\) & 0.20 & \(3.118\pm0.017\) & \(29.72\pm0.26\) & \(7.026\pm0.052\) \\
\bottomrule
\end{tabular}
\endgroup
\captionof{figure}{Tiny ImageNet with 80,000 training images, five seeds per profile. Curves show training and validation performance; the table gives test loss and accuracy at the epoch with the lowest validation loss, together with test loss at the final epoch. All profiles use an initial learning rate of $3\times10^{-4}$.}
\label{fig:tiny-imagenet-80k}
\endgroup\par
\end{minipage}\par\bigskip

\noindent\begin{minipage}{\textwidth}
\subsection{FI-2010}
We train a 12-layer MLP with 256 neurons per hidden layer for 50 epochs without weight decay, using 40,000 training, 10,000 validation and 20,000 test examples. Big-step dropout, chosen by validation loss, reduces test loss by $0.039\pm0.005$ relative to uniform across five seeds, with the uncertainty given by the standard error of the paired differences.

We also compared two linear dropout profiles in a separate Transformer experiment. Those runs have no matching uniform control, so they cannot give an improvement over uniform; their results and training details are available in the repository.
\par\medskip
\begingroup
\centering
\includegraphics[width=0.92\textwidth]{\PaperRoot/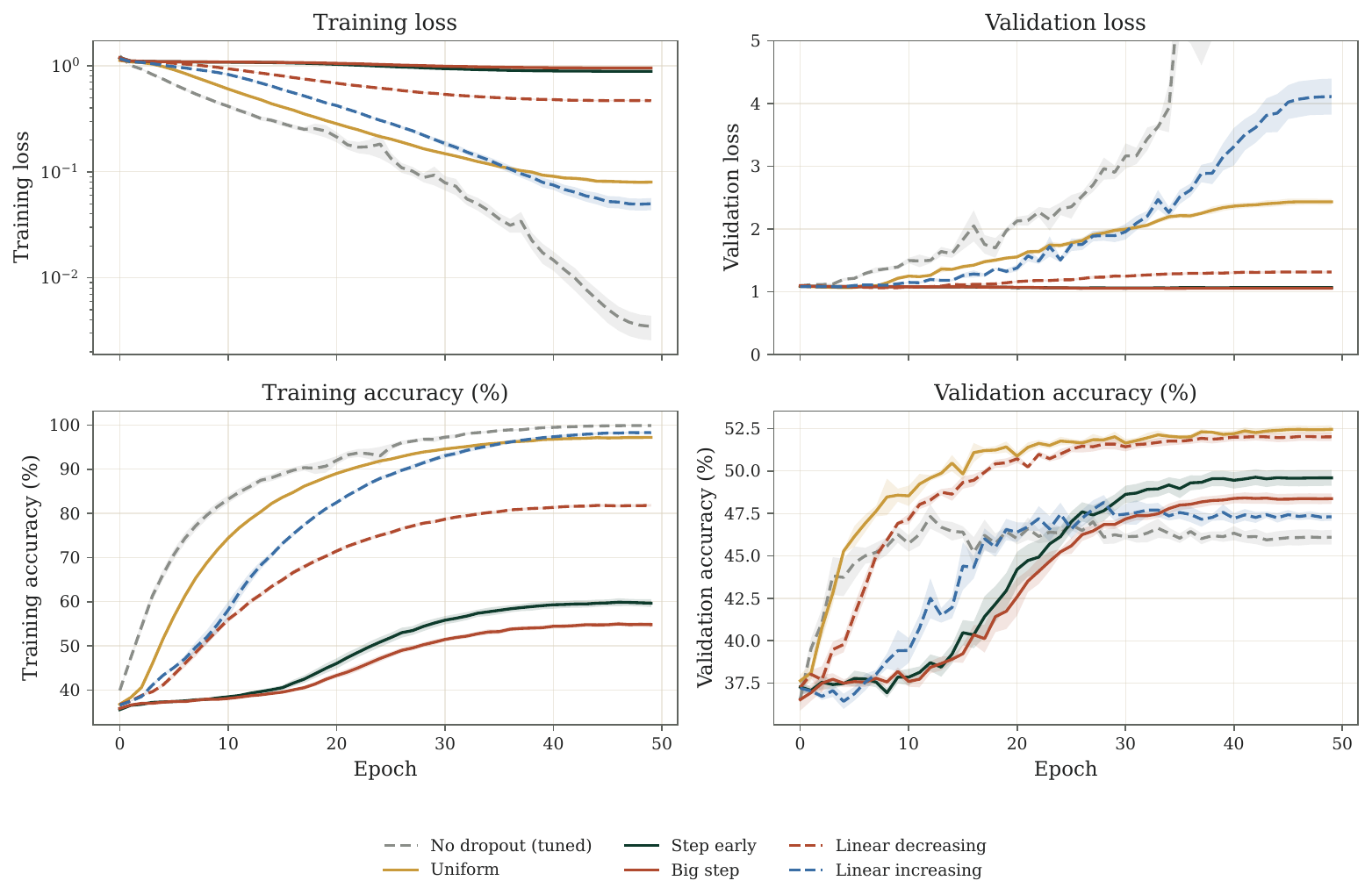}
\par\medskip
\begingroup\small
\begin{tabular}{@{}lrrrr@{}}
\toprule
Profile & $\bar p$ & LR & Test CE & Acc. (\%) \\
\midrule

No dropout & 0.00 & 3e-03 & \(1.118\pm0.017\) & \(42.19\pm2.20\) \\
Uniform & 0.05 & 1e-04 & \(1.093\pm0.004\) & \(47.70\pm0.37\) \\
Step early & 0.10 & 3e-05 & \(1.061\pm0.004\) & \(48.19\pm0.33\) \\
Big step & 0.10 & 3e-05 & \(1.054\pm0.004\) & \(48.37\pm0.39\) \\
Linear decreasing & 0.05 & 3e-05 & \(1.074\pm0.008\) & \(47.48\pm0.27\) \\
Linear increasing & 0.15 & 1e-03 & \(1.115\pm0.011\) & \(37.62\pm0.39\) \\
\bottomrule
\end{tabular}
\endgroup
\captionof{figure}{FI-2010 benchmark MLP: learning curves and test results for five seeds, including the tuned no-dropout control. The no-dropout validation loss continues above the plotted limit of 5.}
\label{fig:fi2010-mlp}
\endgroup\par
\end{minipage}\par\bigskip

\noindent\begin{minipage}{\textwidth}
\subsection{Standalone FI-2010 vanilla MLPs}
These plain feedforward ReLU MLPs flatten a $128\times40$ input window and use day 6 for training, day 7 for validation, and days 8--10 for test. Each architecture trains for 100 epochs with five paired confirmation seeds (100--104), zero weight decay, and mean dropout $0.10$ for nonzero profiles. Weights have variance $1.98/\mathrm{fan\_in}$ and biases variance $0.02$. We use validation loss with uniform dropout to choose one learning rate per architecture, then hold it fixed across profiles, including no dropout. These runs are separate from the benchmark comparison above.

Test CE, accuracy and macro-F1 are evaluated once, at the epoch with the lowest validation loss. The saved results contain means and SEMs, but no validation losses, individual seeds' test results or test curves, so we cannot report final test loss or its minimum during training. Table~\ref{tab:fi2010_standalone_comparison} gives the reductions relative to uniform, choosing the nonuniform profile retrospectively by its mean test CE and reporting accuracy for that same profile.
\par\medskip
\begingroup
\centering
\captionof{table}{Improvements over uniform dropout for the standalone FI-2010 MLPs, calculated from five paired seeds per architecture and using the same profile across each row. Test loss and accuracy are measured at the epoch with the lowest validation loss. The saved summaries lack validation losses, so the profile was chosen retrospectively by mean test loss. Loss reductions compare seed means; accuracy gains are percentage points. Final test results were not saved.}
\label{tab:fi2010_standalone_comparison}
\small
\setlength{\tabcolsep}{6pt}
\begin{tabular}{@{}rrlrr@{}}
\toprule
Depth & Width & Profile & \shortstack{Test loss reduction at\\best validation epoch (\%)} & \shortstack{Accuracy gain\\(pp)} \\
\midrule
6 & 256 & Step late & +1.32 & +0.02 \\
6 & 512 & Linear decreasing & +0.62 & +0.01 \\
12 & 256 & Linear decreasing & +0.43 & -0.06 \\
12 & 512 & Step early & +0.81 & +0.02 \\
\bottomrule
\end{tabular}\par
\endgroup\medskip

\par\medskip
\begingroup
\centering
\captionof{table}{Test results for the standalone FI-2010 MLPs (mean $\pm$ SEM, five paired seeds), measured at the epoch with the lowest validation loss. All controls and early/late profiles are shown.}
\label{tab:fi2010_standalone_profiles}
\small
\setlength{\tabcolsep}{3pt}
\begin{tabular}{@{}lrrr@{\hspace{10pt}}rrr@{}}
\toprule
Profile & \multicolumn{3}{c}{Width 256} & \multicolumn{3}{c}{Width 512} \\
\cmidrule(lr){2-4}\cmidrule(lr){5-7}
 & Test CE & Acc. (\%) & Macro-F1 (\%) & Test CE & Acc. (\%) & Macro-F1 (\%) \\

\midrule
\multicolumn{7}{l}{\textit{Depth 6}} \\
No dropout & \(0.8348\pm0.0071\) & \(69.667\pm0.403\) & \(32.96\pm0.34\) & \(0.8327\pm0.0057\) & \(69.838\pm0.392\) & \(32.63\pm1.27\) \\
Uniform & \(0.8338\pm0.0035\) & \(70.830\pm0.030\) & \(31.74\pm0.31\) & \(0.8318\pm0.0024\) & \(70.854\pm0.020\) & \(31.27\pm0.70\) \\
Step early & \(0.8332\pm0.0024\) & \(70.851\pm0.027\) & \(30.25\pm0.97\) & \(0.8341\pm0.0044\) & \(70.851\pm0.023\) & \(31.41\pm1.05\) \\
Big step & \(0.8279\pm0.0019\) & \(70.839\pm0.010\) & \(30.44\pm1.17\) & \(0.8330\pm0.0042\) & \(70.853\pm0.007\) & \(30.91\pm1.27\) \\
Linear decreasing & \(0.8312\pm0.0035\) & \(70.851\pm0.011\) & \(28.36\pm0.54\) & \(0.8267\pm0.0071\) & \(70.865\pm0.029\) & \(31.15\pm0.81\) \\
Linear increasing & \(0.8293\pm0.0026\) & \(70.865\pm0.009\) & \(31.45\pm0.56\) & \(0.8342\pm0.0037\) & \(70.866\pm0.005\) & \(31.17\pm0.87\) \\
Step late & \(0.8227\pm0.0016\) & \(70.855\pm0.019\) & \(31.97\pm0.45\) & \(0.8308\pm0.0033\) & \(70.841\pm0.036\) & \(31.58\pm0.76\) \\
\midrule
\multicolumn{7}{l}{\textit{Depth 12}} \\
No dropout & \(0.8260\pm0.0062\) & \(70.884\pm0.024\) & \(29.67\pm0.70\) & \(0.8161\pm0.0034\) & \(70.817\pm0.014\) & \(31.53\pm0.98\) \\
Uniform & \(0.8210\pm0.0029\) & \(70.870\pm0.011\) & \(31.93\pm0.89\) & \(0.8228\pm0.0028\) & \(70.829\pm0.016\) & \(31.39\pm0.93\) \\
Step early & \(0.8232\pm0.0032\) & \(70.856\pm0.013\) & \(32.16\pm0.52\) & \(0.8162\pm0.0049\) & \(70.845\pm0.030\) & \(32.11\pm0.65\) \\
Big step & \(0.8224\pm0.0017\) & \(70.809\pm0.040\) & \(32.43\pm0.29\) & \(0.8220\pm0.0031\) & \(70.826\pm0.019\) & \(32.79\pm0.05\) \\
Linear decreasing & \(0.8175\pm0.0025\) & \(70.814\pm0.035\) & \(32.84\pm0.03\) & \(0.8201\pm0.0027\) & \(70.849\pm0.013\) & \(32.87\pm0.02\) \\
Linear increasing & \(0.8227\pm0.0019\) & \(70.844\pm0.017\) & \(32.76\pm0.06\) & \(0.8175\pm0.0019\) & \(70.853\pm0.016\) & \(32.15\pm0.67\) \\
Step late & \(0.8234\pm0.0038\) & \(70.805\pm0.020\) & \(32.78\pm0.09\) & \(0.8238\pm0.0017\) & \(70.835\pm0.024\) & \(31.85\pm1.05\) \\
\bottomrule
\end{tabular}\par
\endgroup\par\medskip
The lowest nonuniform mean test CE occurs for step late at depth 6, width 256, and for linear decreasing at depth 6, width 512 and depth 12, width 256. At depth 12, width 512, no dropout is slightly better than the best nonuniform profile (0.816111 versus 0.816155). The profiles selected for Table~\ref{tab:fi2010_standalone_comparison} exclude no dropout; these results do not establish that nonuniform dropout beats every control.
\end{minipage}\par\bigskip

\noindent\begin{minipage}{\textwidth}
\subsection{CIFAR-10 profile-geometry pilot}
This small experiment compares validation performance for ten fixed profiles, including exact early/late reversals, at mean probability $\bar p=0.10$, with seeds $0,1,2$ (30 runs). Six-layer width-256 ReLU MLPs train on 4,000 class-balanced examples with 1,000 validation examples (split seed 20260812), using AdamW, LR $10^{-4}$, weight decay $10^{-7}$, batch size 75, and 75 epochs of the benchmark cosine schedule without clipping. Images use fixed channel means $(0.4914,0.4822,0.4465)$ and standard deviations $(0.2470,0.2435,0.2616)$, without augmentation. Power profiles use layer-cell centers and cap $0.20$. The manifest records the exact arrays, settings and source/split hashes; the code for this pilot is not included in the released benchmark code. We do not tune hyperparameters or evaluate on the test set, so these runs do not enter the main test comparisons.
\par\medskip
\begingroup\centering
\includegraphics[width=0.85\textwidth]{\PaperRoot/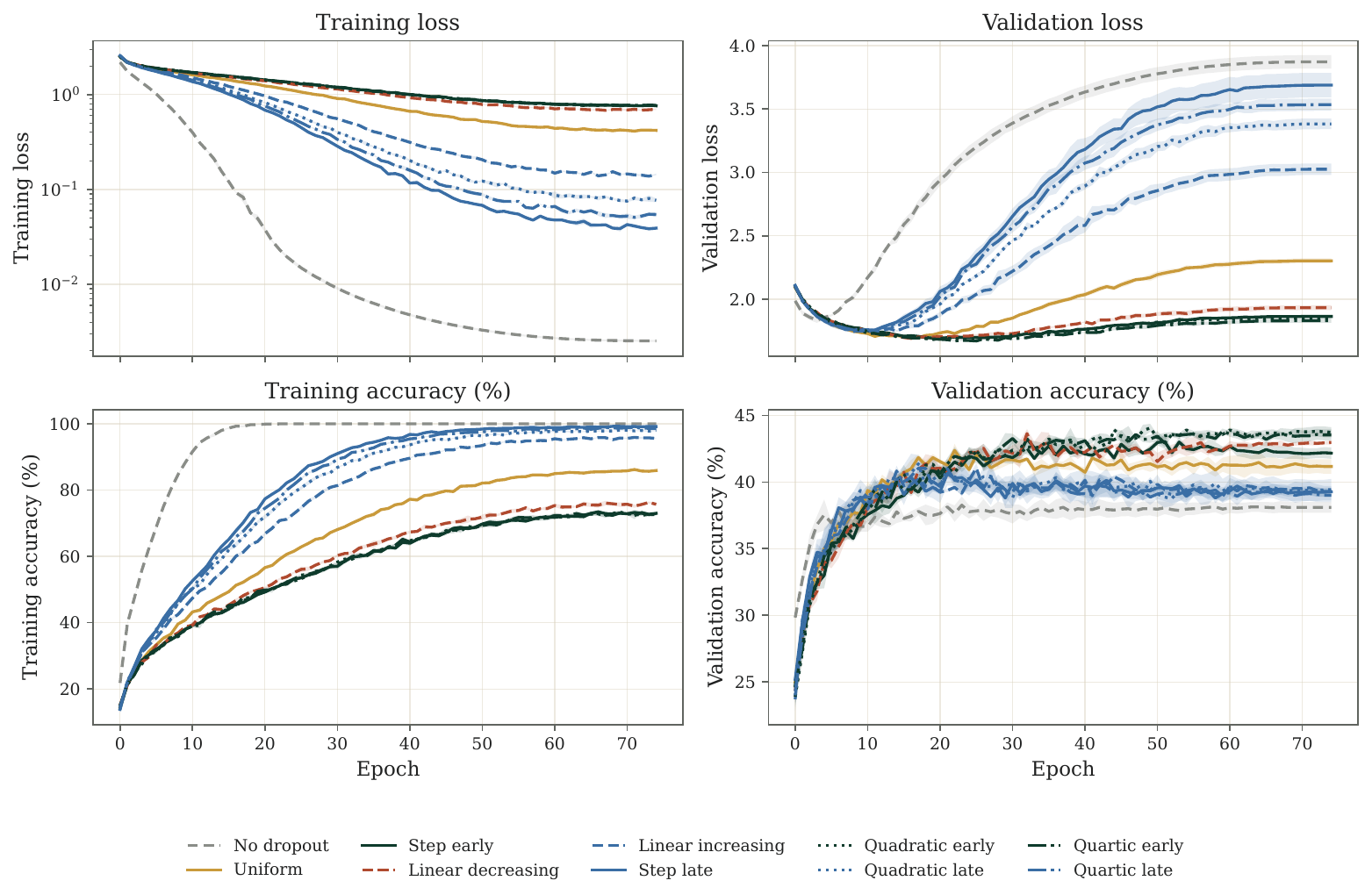}
\captionof{figure}{Different dropout profiles at fixed mean dropout. Training and validation curves show mean $\pm$ SEM over three seeds.}
\label{fig:profile-geometry-pilot}\endgroup\par
\begin{center}\small\begin{tabular}{@{}lrr@{}}\toprule
Profile & Final validation CE & Final validation accuracy (\%) \\
\midrule
No dropout & \(3.872\pm0.053\) & \(38.10\pm0.40\) \\
Uniform & \(2.303\pm0.019\) & \(41.17\pm0.55\) \\
Step early & \(1.864\pm0.009\) & \(42.17\pm0.43\) \\
Linear decreasing & \(1.933\pm0.019\) & \(42.97\pm0.15\) \\
Linear increasing & \(3.025\pm0.045\) & \(39.30\pm0.40\) \\
Step late & \(3.688\pm0.097\) & \(39.27\pm0.30\) \\
Quadratic early & \(1.852\pm0.001\) & \(43.73\pm0.38\) \\
Quadratic late & \(3.382\pm0.042\) & \(39.43\pm0.80\) \\
Quartic early & \(1.831\pm0.016\) & \(43.53\pm0.41\) \\
Quartic late & \(3.534\pm0.048\) & \(39.00\pm0.50\) \\
\bottomrule\end{tabular}\end{center}
\end{minipage}\par\bigskip

\end{document}